\documentclass[10pt,twocolumn,letterpaper]{article}

\usepackage[pagenumbers]{cvpr} 

\definecolor{cvprblue}{rgb}{0.21,0.49,0.74}
\usepackage[pagebackref,breaklinks,colorlinks,allcolors=cvprblue]{hyperref}
\usepackage{wrapfig}
\usepackage{graphicx}
\usepackage{cuted}
\usepackage{multirow}
\usepackage{capt-of}
\usepackage{makecell}
\usepackage{natbib}
\usepackage{amsmath}
\usepackage{amsfonts}
\usepackage{bbold}
\usepackage{array}
\usepackage{algorithm}
\usepackage{algorithmicx}
\usepackage{algpseudocode}

\usepackage{pifont}
\usepackage{siunitx}
\usepackage{booktabs}       
\usepackage{threeparttable} 
\usepackage{array}          
\usepackage[table]{xcolor}  
\usepackage{amssymb}
\usepackage{xcolor}   
\usepackage{titling}

\def\paperID{12237} 
\def\confName{CVPR}
\def\confYear{2026}

\title{\textbf{Dynamic-eDiTor: Training-Free Text-Driven 4D Scene Editing with Multimodal Diffusion Transformer}}
\date{}

\renewcommand{\thefootnote}{\fnsymbol{footnote}}

\author{
Dong In Lee\textsuperscript{1,2}\protect\footnotemark[1] \ \footnotemark[3] \quad
Hyungjun Doh\textsuperscript{1}\protect\footnotemark[1] \quad
Seunggeun Chi\textsuperscript{1} \quad
Runlin Duan\textsuperscript{1} \quad \vspace{0.2em}\\
Sangpil Kim\textsuperscript{2}\protect\footnotemark[2] \quad
Karthik Ramani\textsuperscript{1}\protect\footnotemark[2] \vspace{0.5em}\\
\textsuperscript{1}Purdue University \quad
\textsuperscript{2}Korea University
}

\begin{document}
\maketitle

\footnotetext[1]{Co-first authors.}
\footnotetext[2]{Co-corresponding authors.}
\footnotetext[3]{Work done at Purdue University as a visiting scholar.}

\renewcommand{\thefootnote}{\arabic{footnote}}

\begin{strip}\centering
\vspace{-6em}
\includegraphics[width=\textwidth]{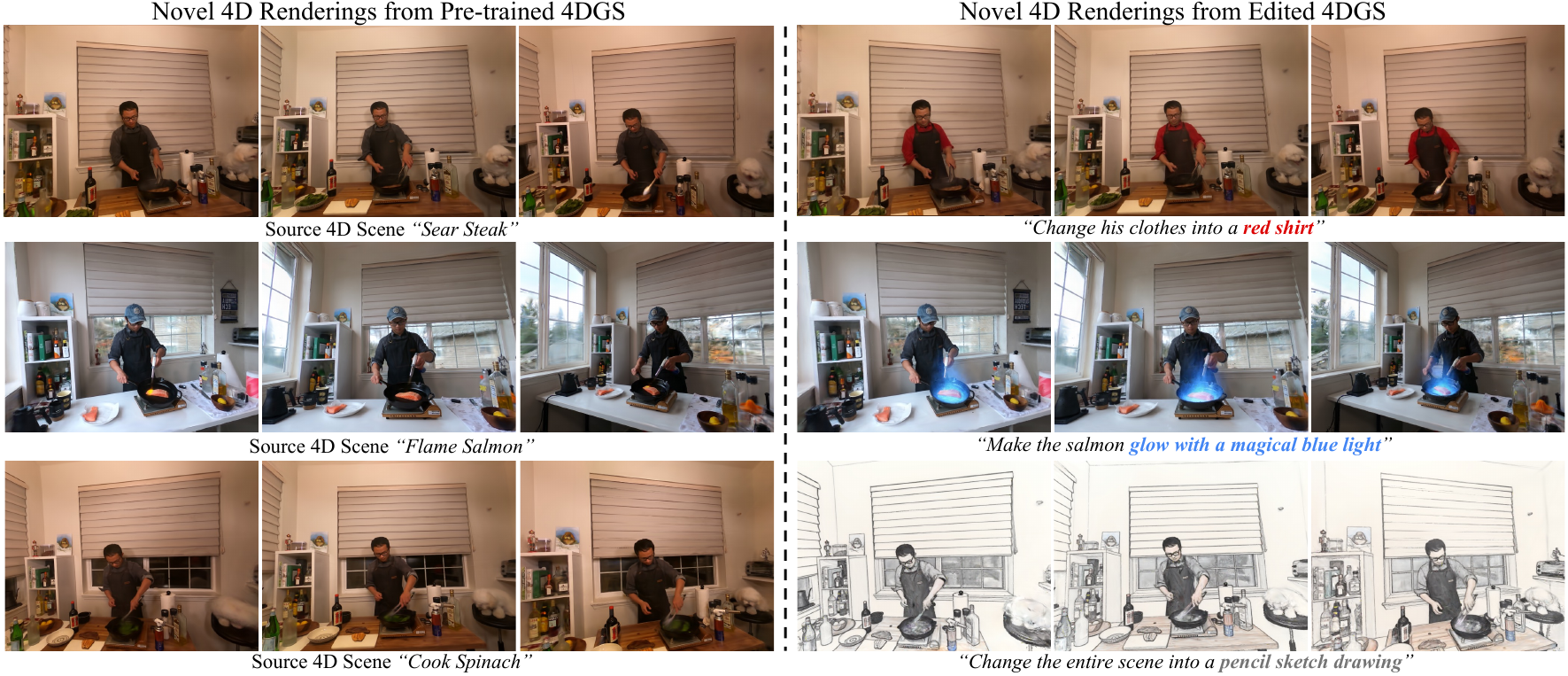}
\vspace{-2.em}
\captionof{figure}
{We propose \textbf{Dynamic-eDiTor} enables flexible and high-quality editing of pre-trained 4D Gaussian Splatting~\cite{wu20244d} models leveraging Multimodal Diffusion Transformer~\cite{esser2024scaling, wu2025qwen} guided solely by text instructions. Through its design focused on both multi-view and temporal consistency, our approach demonstrates robust performance, producing realistic and fine-grained 4D scene manipulation.
}
\vspace{-0.5em}

\label{fig:teaser}
\end{strip}

\begin{abstract}

Recent progress in 4D representations, such as Dynamic NeRF and 4D Gaussian Splatting (4DGS), has enabled dynamic 4D scene reconstruction. However, text-driven 4D scene editing remains under-explored due to the challenge of ensuring both multi-view and temporal consistency across space and time during editing.
Existing studies rely on 2D diffusion models that edit frames independently, often causing motion distortion, geometric drift, and incomplete editing. We introduce Dynamic-eDiTor, a training-free text-driven 4D editing framework leveraging Multimodal Diffusion Transformer (MM-DiT) and 4DGS. This mechanism consists of Spatio-Temporal Sub-Grid Attention (STGA) for locally consistent cross-view and temporal fusion, and Context Token Propagation (CTP) for global propagation via token inheritance and optical-flow-guided token replacement. Together, these components allow Dynamic-eDiTor to perform seamless, globally consistent multi-view video without additional training and directly optimize pre-trained source 4DGS.
Extensive experiments on multi-view video dataset DyNeRF demonstrate that our method achieves superior editing fidelity and both multi-view and temporal consistency prior approaches. Project page for results and code:: \url{https://di-lee.github.io/dynamic-eDiTor/}


\end{abstract}    
\vspace{-2em}
\section{Introduction}
\label{sec:intro}

Recent advances in 3D representations, such as Neural Radiance Field (NeRF)~\cite{mildenhall2021nerf} and 3D Gaussian Splatting (3DGS)~\cite{kerbl20233d}, have achieved significant progress in photo-realistic 3D reconstruction of real-world scenes. More recently, 4D representations such as Dynamic NeRF~\cite{pumarola2021d} and 4D Gaussian Splatting (4DGS)~\cite{wu20244d} extend 3D representations into the time domain, enabling spatio-temporally coherent reconstruction.  
However, text-driven 4D scene editing remains under-explored, primarily due to the difficulty of maintaining both multi-view and temporal consistency across space and time during editing.


In this work, we focus on the multi-view video setting of 4DGS, which provides richer viewpoint coverage but further amplifies the difficulty of achieving both multi-view and temporal consistency during editing. While 3D editing primarily focuses on multi-view consistency, 4D editing introduces the further challenge of ensuring both multi-view and temporal consistency across viewpoints and time.




Recent studies~\cite{mou2024instruct, he2025ctrl, kwon2025efficient} have attempted 4D scene editing by combining 2D diffusion models with 4D representations~\cite{song2023nerfplayer, wu20244d}. However, these methods typically perform frame-wise editing or require per-scene finetuning of the 2D diffusion model~\cite{brooks2023instructpix2pix}, lacking a unified mechanism to jointly process information across views and time. Consequently, they struggle with non-rigid content manipulation and are often limited to style-oriented edits, leading to motion distortions, geometric drift, and incomplete editing results.


To address these limitations, we propose \textbf{Dynamic-eDiTor}, a novel training-free, text-driven 4D editing framework that leverages Multimodal Diffusion Transformer (MM-DiT)~\cite{esser2024scaling, wu2025qwen} and 4DGS. 
Our goal is to maintain globally coherent motion and geometry while enabling flexible, semantically grounded edits.
To this end, we propose Grid-based Spatio-Temporal Propagation, which represents the entire multi-view video as a unified camera-time grid and jointly aggregates spatial and temporal information and propagates the fused features across throughout the grid.





As its foundation, we introduce Spatio-Temporal Sub-Grid Attention (STGA), which extends MM-DiT’s dual-stream self-attention to operate on localized spatio-temporal sub-grids. By jointly attending to adjacent viewpoints and neighboring time steps, STGA enables coherent local feature fusion without additional training. We additionally identify a vital layer range in MM-DiT where incorporating STGA yields the strongest improvements in multi-view and temporal consistency.



To ensure that the fused information is globally propagated throughout the multi-view video, we further introduce Context Token Propagation (CTP), an explicit propagation mechanism that transfers fused tokens along a structured traversal path over the entire multi-view video. Tokens in overlapping regions are fully inherited, while non-overlapping temporal regions are updated through flow-guided token warping using optical flow~\cite{teed2020raft}. This unified propagation strategy ensures coherent feature flow across views and time, reinforcing multi-view and temporal consistency and enabling stable, high-fidelity 4D optimization.

Finally, the edited frames are directly used to optimize the pre-trained 4DGS without the Iterative Dataset Update (IDU)~\cite{mou2024instruct, he2025ctrl}, resulting in globally consistent 4D content that faithfully reflects the desired edits.



We validate Dynamic-eDiTor on the multi-view video dataset DyNeRF~\cite{li2022neural}, achieving superior editing fidelity, temporal smoothness, and robustness compared to state-of-the-art methods. Our key contributions are as follows:

\begin{itemize}

\item We present Dynamic-eDiTor, a novel training-free, text-driven 4D editing framework that leverages MM-DiT~\cite{esser2024scaling, wu2025qwen} and 4DGS~\cite{wu20244d} to enable spatially and temporally consistent dynamic 4D scene editing.

\item We propose Spatio-Temporal Sub-Grid Attention (STGA), which jointly attends across adjacent viewpoints and neighboring time steps to integrate spatial and temporal features on a vital layer range in MM-DiT.

\item We introduce Context Token Propagation (CTP), an explicit propagation mechanism that distributes fused spatio-temporal information across the entire multi-view video by inheriting tokens in overlapping regions and replacing non-overlapping regions via flow-based warping.


\item Through extensive qualitative and quantitative experiments, we demonstrate that Dynamic-eDiTor significantly outperforms existing methods in 4D editing fidelity, spatio-temporal stability, and robustness.

\end{itemize}

\begin{figure*}[t]
    \vspace{-2.5em}
    \centering
    \includegraphics[width=\linewidth]{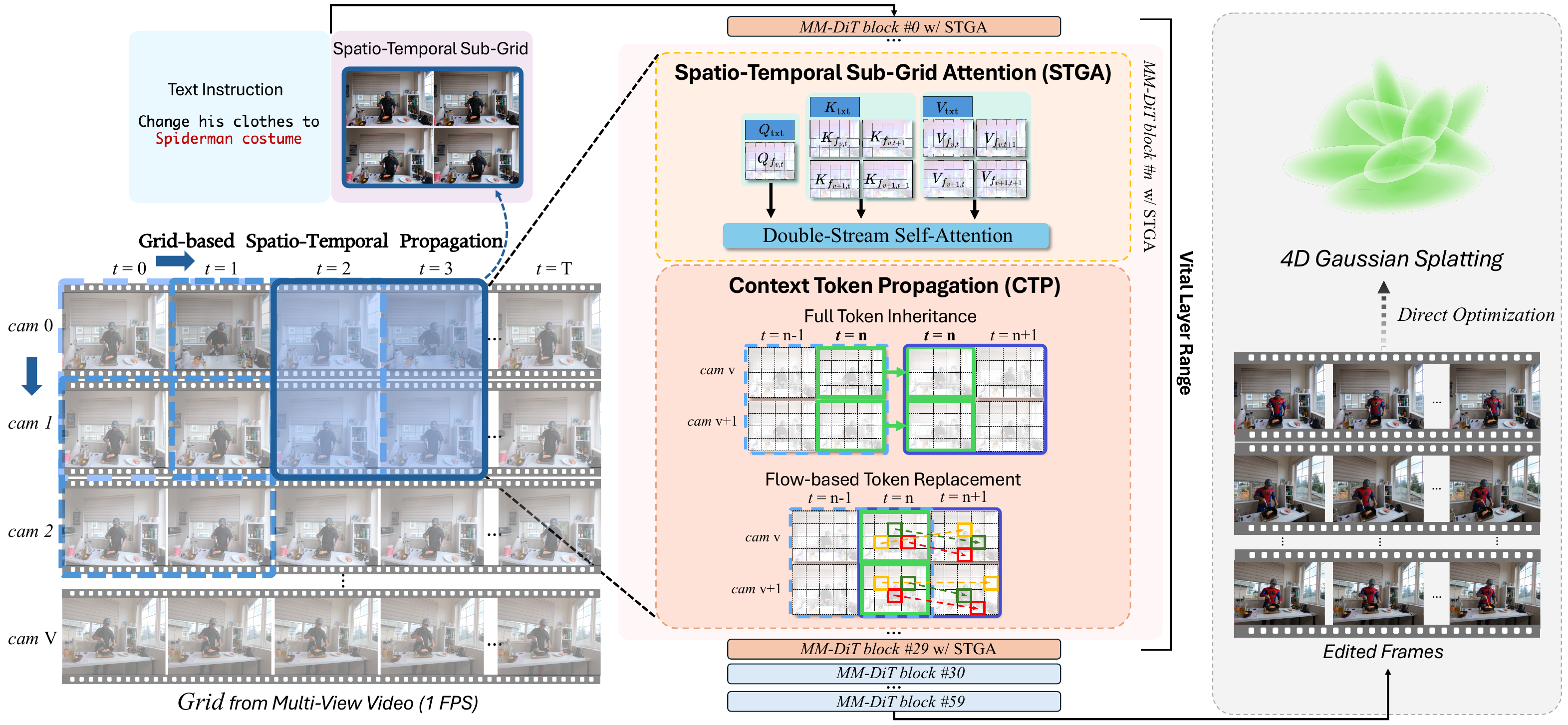}
    \vspace{-2.0em}
    \caption{
    \textbf{Dynamic-eDiTor Overview.} We represent the multi-view video as a unified camera–time grid. Dynamic-eDiTor combines Spatio-Temporal Sub-Grid Attention (STGA), which performs locally coherent cross-view and temporal fusion within each sub-grid, with Context Token Propagation (CTP), which globally propagates the aggregated features across the grid via Full Token Inheritance and Flow-guided Token Replacement for robust spatio-temporal consistency enforcement. Together, these modules enable seamless, globally consistent multi-view video editing without additional training, while directly optimizing the pre-trained 4DGS.
    }
    \label{fig:overview}
    \vspace{-1.5em}
\end{figure*}
\vspace{-0.5em}
\section{Related Work}
\label{sec:related_work}

\subsection{2D Editing} 


2D diffusion models have demonstrated remarkable generalization and controllability for text-guided image editing.
U-Net-based~\cite{ronneberger2015u} models such as Prompt-to-Prompt~\cite{hertz2022prompt}, SDEdit~\cite{meng2021sdedit}, and InstructPix2Pix~\cite{brooks2023instructpix2pix} enable text-guided image manipulation via controlled denoising.
More recently, Diffusion Transformers (DiT)~\cite{peebles2023scalable} replace the U-Net backbone with a Transformer architecture~\cite{vaswani2017attention, dosovitskiy2020image}, offering improved scalability and visual coherence. This approach has evolved into multimodal variants such as Multimodal Diffusion Transformer (MM-DiT)~\cite{esser2024scaling}. MM-DiT employs a dual-stream architecture, processing text and image tokens in parallel streams fused via joint attention. 
Building on this trend, MM-DiT-based editors such as FLUX~\cite{flux2024, labs2025flux1kontextflowmatching}, HiDream~\cite{cai2025hidream}, and Qwen-Image-Edit~\cite{wu2025qwen} achieve precise instruction-driven image editing.
Leveraging the MM-DiT, our Dynamic-eDiTor extends a MM-DiT-based image editor to maintain consistency across time and viewpoints within a unified 4D framework.

\subsection{3D Scene Editing}
Neural Radiance Fields (NeRF)~\cite{mildenhall2021nerf} and 3D Gaussian Splatting (3DGS)~\cite{kerbl20233d} have enabled high-fidelity 3D reconstruction and inspired extensive research on 3D scene editing.
Instruct-NeRF2NeRF~\cite{haque2023instruct} introduces the Iterative Dataset Update (IDU) that edits the rendered image using a 2D diffusion model~\cite{brooks2023instructpix2pix} while optimizing the underlying 3D model, NeRF. GaussianEditor~\cite{wang2024gaussianeditor} adopts IDU on 3DGS and explicitly controls 3D Gaussians. Recently, EditSplat~\cite{lee2025editsplat} achieves high-fidelity 3D edits by integrating multi-view information into the diffusion process and explicitly pruning 3DGS. Despite these advances, existing 3D editing methods~\cite{dong2023vica, wang2024view, chen2024dge, wu2024gaussctrl, zhu2025coreeditor, zhao2025tinker, karim2024free} primarily focus on enforcing multi-view consistency within static scenes, and cannot handle temporal dynamics across frames.


\subsection{4D Scene Editing} 
Extending 3D scene editing to 4D representations introduces the additional challenge of maintaining temporal consistency while preserving multi-view coherence.
Instruct 4D-to-4D~\cite{mou2024instruct} first applied diffusion-based 2D editing to sequentially rendered frames while optimizing the underlying NeRF-based 4D model, while 4D-Editor~\cite{jiang20254d} incorporates spatial segmentation and motion-aware propagation for object-level editing. Control4D~\cite{shao2024control4d} enables 4D portrait editing by distilling the knowledge from a 2D diffusion into a 4D GAN~\cite{goodfellow2014generative} generator.
CTRL-D~\cite{he2025ctrl} finetunes InstructPix2Pix~\cite{brooks2023instructpix2pix} with prior-preservation loss~\cite{ruiz2023dreambooth} per scene for consistent 2D edits and iteratively optimizes 4DGS. 
Instruct4DGS~\cite{kwon2025efficient} edits canonical Gaussians first and employs score-distillation-based~\cite{poole2022dreamfusion} refinement for temporal smoothness. While these methods demonstrate notable progress, they still edit frames independently without simultaneously processing information across views and time, often causing motion or geometric distortion. In contrast, our Dynamic-eDiTor achieves consistent 4D editing by jointly editing cross-view and temporal frames, and by propagating these context tokens to the entire multi-view video.


\section{Preliminary}

\subsection{4D Scene Representation}
3D Gaussian Splatting (3DGS)~\cite{kerbl20233d} represents a scene as a set of anisotropic Gaussian primitives 
\(\mathcal{G} = \{ (\mu_i, \Sigma_i, c_i, \alpha_i) \}_{i=1}^{N}\), 
each defined by its position, covariance, color, and opacity. Rendering is performed via differentiable alpha compositing. 
To model dynamic scenes, 4D Gaussian Splatting (4DGS)~\cite{wu20244d} extends this representation with a deformation field that maps canonical Gaussians to their deformed states over time. The field predicts offsets for position, rotation, and scale using MLPs \(\phi_x\), \(\phi_r\), and \(\phi_s\): \(\Delta x = \phi_x(z)\), \(\Delta r = \phi_r(z)\), and \(\Delta s = \phi_s(z)\), where \(z\) is a temporal feature encoding the dynamic state of the scene. 
The final deformed Gaussian parameters are obtained as: 
{
\setlength{\abovedisplayskip}{5pt}
\setlength{\belowdisplayskip}{5pt}
\setlength{\abovedisplayshortskip}{5pt}
\setlength{\belowdisplayshortskip}{5pt}
\begin{equation}
(x', r', s') = (x + \Delta x,\, r + \Delta r,\, s + \Delta s), 
\end{equation}
}
yielding the time-varying Gaussian set \(\mathcal{G}'\).
\vspace{-0.3em}
\section{Method}

We propose a novel training-free 4D editing framework, \textbf{Dynamic-eDiTor}, carefully designed to achieve spatially and temporally consistent 4D scene editing leveraging Multimodal Diffusion Transformer (MM-DiT)~\cite{esser2024scaling, wu2025qwen}.  Initially, our approach edits source multi-view video frames at 1 FPS corresponding to a pre-trained 4D Gaussian Splatting (4DGS)~\cite{wu20244d}, ensuring both multi-view and temporal consistency. 
The edited frames are then used to directly optimize the underlying pre-trained 4DGS representation.


\subsection{Grid-based Spatio-Temporal Propagation} 
To ensure both multi-view and temporal consistency in multi-view video editing, we introduce Grid-based Spatio-Temporal Propagation, which enables coherent feature flow across the entire scene through two components:  (1) Spatio-Temporal Sub-Grid Attention (STGA) for local fusion across adjacent views and neighboring time steps and (2) Context Token Propagation (CTP) for globally propagating the fused information through the entire multi-view video.

We begin by representing all multi-view video frames as a unified camera--time grid:
{
\setlength{\abovedisplayskip}{5pt}
\setlength{\belowdisplayskip}{5pt}
\setlength{\abovedisplayshortskip}{5pt}
\setlength{\belowdisplayshortskip}{5pt}
\begin{equation}
Grid = \{f_{v,t} \mid v \in [0, \dots, V],\ t \in [0, 
\dots, T] \},
\label{eq:Grid}
\end{equation}
}
where \(f_{v,t}\) is the frame captured by viewpoint \(v\) at time index \(t\). The Grid's rows correspond to different viewpoints and  columns represent temporally aligned frames.

To enable localized spatio-temporal fusion, we partition the \(Grid\) into overlapping 
$2\times2$ sub-grid \(\mathcal{S}_{v,t}\) at position \((v,t)\) on \(Grid\) defined as:
{
\setlength{\abovedisplayskip}{5pt}
\setlength{\belowdisplayskip}{5pt}
\setlength{\abovedisplayshortskip}{5pt}
\setlength{\belowdisplayshortskip}{5pt}
\begin{equation}
\mathcal{S}_{v,t} = \{ f_{v,t},\ f_{v+1,t},\ f_{v,t+1},\ f_{v+1,t+1} \},
\label{eq:subgrid}
\end{equation}
}
each covering adjacent views and neighboring time steps.
These sub-grids serve as the atomic processing units for STGA, allowing each local region to aggregate and share information across both the view and temporal axes.

To propagate information across the entire $V \times T$ \(Grid\), 
we process the sub-grids sequentially using an asymmetric sliding pattern. 
We first sweep vertically along the spatial axis at $t{=}0$ to establish multi-view alignment, 
and then slide horizontally along the temporal axis to propagate consistency over time. 
The induced overlaps between neighboring sub-grids provide the structural linkage for STGA and CTP to effectively distribute information, enforcing globally coherent spatio-temporal editing.




\subsubsection{Spatio-Temporal Sub-Grid Attention (STGA)}
Grid-based Spatio-Temporal Propagation's foundation is the fusion of information within local neighborhoods. We propose Spatio-Temporal Sub-Grid Attention (STGA), which extends the dual-stream self-attention mechanism in MM-DiT architecture~\cite{esser2024scaling} to jointly attend adjacent views and temporally neighboring frames.

Instead of processing each frame's feature independently as in standard MM-DiT, STGA operates on a local sub-grid \(\mathcal{S}_{v,t}\), enabling each frame to aggregate features from its cross-view and temporal neighbors. Each sub-grid contains four frames, and every frame \(f_i \in \mathcal{S}_{v,t}\) is sequentially processed as a query in turn. 
Following MM-DiT's dual-stream attention design, the attention calculation involves two parts--the text stream and image-feature stream. For given frame \(f_i\), we use its image query \(Q_{f_i}\). We then extend the image-feature stream by concatenating all frame-level features within the sub-grid \(\mathcal{S}_{v,t}\) to construct joint key and value sets \(K_{\mathcal{S}_{v,t}}\) and \(V_{\mathcal{S}_{v,t}}\): 
{
\setlength{\abovedisplayskip}{5pt}
\setlength{\belowdisplayskip}{5pt}
\setlength{\abovedisplayshortskip}{5pt}
\setlength{\belowdisplayshortskip}{5pt}
\begin{equation}
\begin{aligned}
K_{\mathcal{S}_{v,t}} &= [K_{f_{v,t}}, K_{f_{v+1,t}}, K_{f_{v,t+1}}, K_{f_{v+1,t+1}}], \\
V_{\mathcal{S}_{v,t}} &= [V_{f_{v,t}}, V_{f_{v+1,t}}, V_{f_{v,t+1}}, V_{f_{v+1,t+1}}].
\end{aligned}
\label{eq:qkv}
\end{equation}
}
The STGA for each frame \(f_i\) integrates the text stream \(Q_{\text{txt}}, K_{\text{txt}}, V_{\text{txt}}\) with our modified, spatio-temporal image stream. We then apply Rotary Position Embeddings (RoPE) to the image queries \(Q_{f_i}, K_{\mathcal{S}_{v,t}}, V_{\mathcal{S}_{v,t}}\) to provide positional information before the softmax operation:
{
\setlength{\abovedisplayskip}{5pt}
\setlength{\belowdisplayskip}{5pt}
\setlength{\abovedisplayshortskip}{5pt}
\setlength{\belowdisplayshortskip}{5pt}
\begin{equation}
\begin{aligned}
\mathrm{STGA}&(\mathcal{S}_{v,t}) 
= \mathrm{softmax}\Big(
    [Q_{\text{txt}},\, \mathrm{RoPE}(Q_{f_{v,t}})] \: \cdot \\
&    [K_{\text{txt}},\, \mathrm{RoPE}(K_{\mathcal{S}_{v,t}})]^{\top} / \sqrt{d_k}
\Big)
\cdot [V_{\text{txt}},\, V_{\mathcal{S}_{v,t}}],
\label{eq:joint_attn}
\end{aligned}
\end{equation}
}
where \(d_k\) denotes the dimensionality of the key vectors. 

STGA encourages cross-view and temporal coherence, forming the foundation for globally consistent multi-view video editing. Unlike previous temporal-only extensions of self-attention~\cite{wu2023tune, geyer2023tokenflow}, our STGA enables each frame query to attend both spatially adjacent views and temporally neighboring frames within its saptio-temporal patch. While STGA operates locally within each sub-grid, the overlapping sliding 
pattern naturally leads to implicit propagation of fused information across 
adjacent sub-grids.


\begin{figure}[t]
    \centering
    \includegraphics[width=\linewidth]{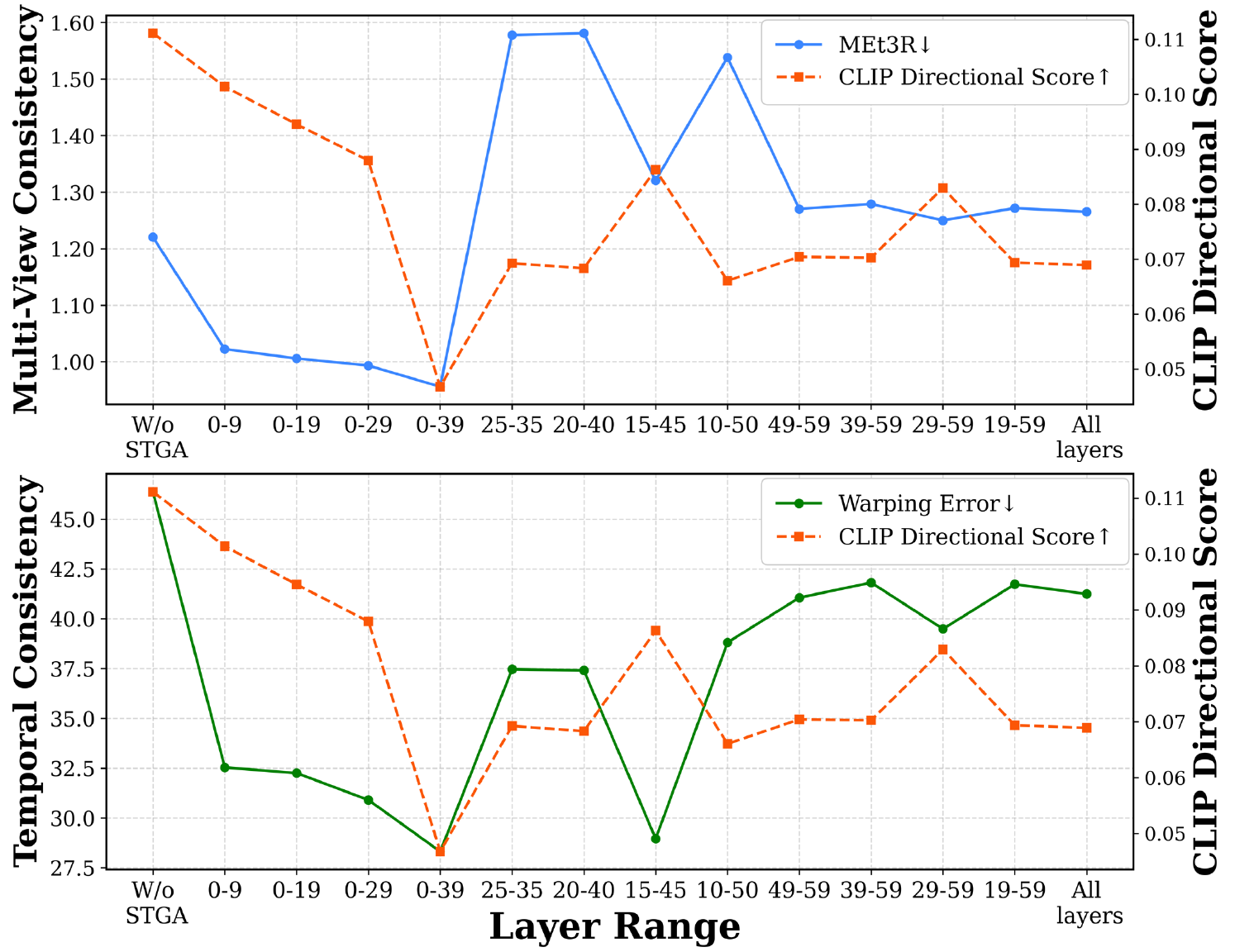}
    \vspace{-2.em}
    \caption{\textbf{Vital Layer Range Analysis.} We analyze the impact of applying Spatio-Temporal Sub-Grid Attention (STGA) across different layer ranges in MM-DiT~\cite{wu2025qwen,esser2024scaling} during the multi-view video editing process. Performance is evaluated by temporal consistency (\textit{Warping Error}~\cite{lai2018learning}), multi-view consistency (\textit{MEt3R}~\cite{asim2025met3r}), and editing fidelity (\textit{CLIP Text-Image Directional Similarity~\cite{radford2021learning}}). 
    Applying STGA to the early $\sim$30 layers provides the best trade-off between consistency and editing fidelity.}
    \label{fig:layer_analysis}
    \vspace{-1.5em}
\end{figure}

\begin{figure*}[t]
    \vspace{-2.2em}
    \centering
    \includegraphics[width=\linewidth]{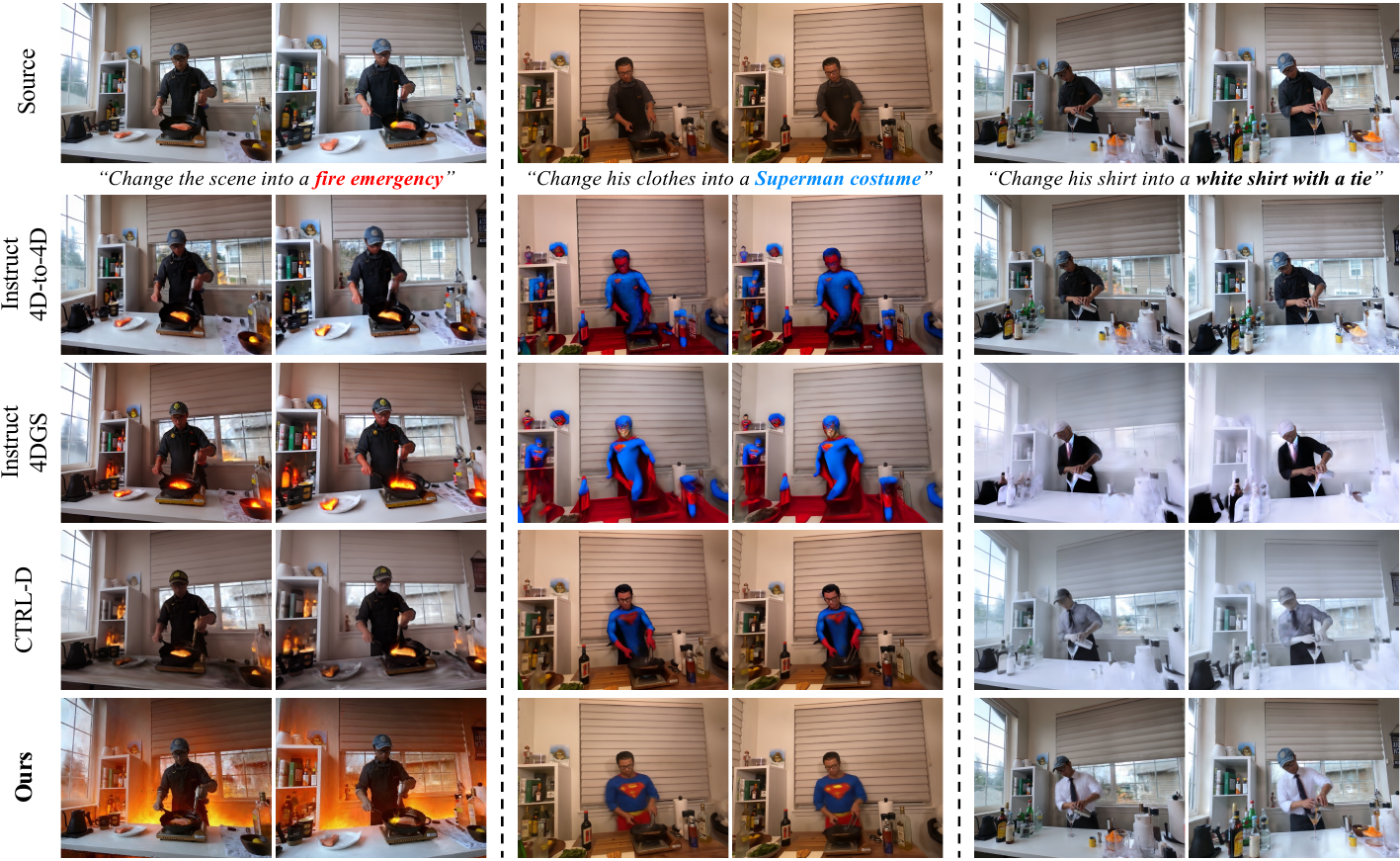}
    \vspace{-2.em}
    \caption{
    \textbf{Qualitative Comparison.} Dynamic-eDiTor enables more robust non-rigid content manipulation and achieves more complete edits of the 4D scene. The top-row displays the original rendered frames, while the following rows show the edited 4DGS renderings produced by each baseline. Our method (bottom-row) outperforms all baselines in both text alignment and overall editing fidelity, while maintaining strong temporal and spatial consistency.
    }
    \vspace{-1.5em}
    \label{fig:comparison}
\end{figure*}

\vspace{-1.em}
\label{vital_layer}
\paragraph{Vital Layer Range Selection.}
As illustrated in~\cref{fig:ablation_vital}, applying STGA to all self-attention layers of MM-DiT~\cite{wu2025qwen, esser2024scaling} leads to visual artifacts, as the STGA tends to over-attend within the local spatio-temporal patch, resulting in texture repetition and view-dependent inconsistencies~\cite{feng2025personalize}.
While prior studies~\cite{gao2025ditvr, avrahami2025stable, kim2025tv, wei2025freeflux} analyze layer importance in DiT-based models~\cite{peebles2023scalable} by ranking individual layers. 
Instead, we investigate applying STGA across continuous layer ranges to capture this cumulative effect. We empirically observe that there is a vital layer range to apply STGA for effectively enforcing multi-view and temporal consistency while alleviating editing-quality degradation. 
As shown in~\cref{fig:layer_analysis}, applying STGA to the first 30 layers achieves the best trade-off between consistency and fidelity, providing significant improvements in both multi-view and temporal coherence while maintaining editing quality.


\subsubsection{Context Token Propagation (CTP)}
While STGA achieves local cross-view and temporal coherence by jointly attending adjacent views and neighboring frames within the sub-grid, Context-Aware Propagation ensures this coherence is globally distributed across entire \(V \times T\ Grid\).
As the sub-grid slides along the defined traversal path, we introduce Context Token Propagation (CTP), which explicitly propagates context information. This ensures that the coherent feature representations that contain spatial and temporal information computed by STGA in the previous sub-grid \(S_{prev}\) are injected into the current sub-grid \(S_{curr}\), ensuring coherent feature flow, enforcing global consistency and preventing information loss.






\begin{table*}[htbp]
\footnotesize
    \centering
    \renewcommand\arraystretch{1.0}
    \setlength{\tabcolsep}{0.8mm}{
    \begin{threeparttable}
    \begin{tabular}{cccccccccccc}
\toprule
    & \multicolumn{2}{c}{\textbf{Editing Fidelity}} & \multicolumn{6}{c}{\textbf{User Study}} & \multicolumn{3}{c}{\textbf{Reconstruction Fidelity}} \\
    \cmidrule(lr){2-3} \cmidrule(lr){4-9} \cmidrule(lr){10-12} 
    
    \makecell{\textbf{Method}}
    & \makecell{\footnotesize CLIP$_{\text{dir}}$ $\uparrow$} 
    & \makecell{\footnotesize CLIP$_{\text{sim}}$ $\uparrow$} 
    & \makecell{\footnotesize Overall \\ Quality $\scriptscriptstyle(\%)$} 
    & \makecell{\footnotesize Motion \\ Consist. $\scriptscriptstyle(\%)$} 
    & \makecell{\footnotesize Temporal \\ Consist. $\scriptscriptstyle(\%)$} 
    & \makecell{\footnotesize Multi-view \\ Consist. $\scriptscriptstyle(\%)$} 
    & \makecell{\footnotesize Prompt \\ Align. $\scriptscriptstyle(\%)$} 
    & \makecell{\footnotesize Identity \\ Preserv. $\scriptscriptstyle(\%)$} 
    & \makecell{\footnotesize PSNR $\uparrow$} 
    & \makecell{\footnotesize SSIM $\uparrow$} 
    & \makecell{\footnotesize LPIPS $\downarrow$} \\
    \midrule
        \text{Instruct4D-to-4D~\cite{mou2024instruct}} 
        & 0.1077 & 0.6308 
        & \underline{27.57} & \underline{27.72} & \underline{28.00} & \underline{27.19} & \underline{22.14} & \underline{27.48}
        & 21.86 & 0.6978 & 0.2145 \\
        
        \text{Instruct-4DGS~\cite{kwon2025efficient}} 
        & \underline{0.1501} & \underline{0.6342} 
        & 10.48 & 10.52 & 11.05 & 10.48 & 9.29 & 11.05
        & 20.62 & 0.6252 & 0.2869 \\
        
        \text{CTRL-D~\cite{he2025ctrl}} 
        & 0.1498 & 0.6141 
        & 13.00 & 14.62 & 15.57 & 14.14 & 11.71 & 13.95
        & \textbf{31.06} & \textbf{0.8498} & \textbf{0.0970} \\
        
        \cmidrule(r){1-1} \cmidrule(l){2-3} \cmidrule(l){4-9} \cmidrule(l){10-12} 
        \textbf{Ours} 
        & \textbf{0.1849} & \textbf{0.6397} 
        & \textbf{48.95} & \textbf{47.14} & \textbf{45.38} & \textbf{48.19} & \textbf{56.86} & \textbf{47.52}
        & \underline{29.25} & \underline{0.8064} & \underline{0.1006} \\
        \bottomrule
    \end{tabular}
    \end{threeparttable}}
    \vspace{-1.em}
    \caption{\textbf{Quantitative Comparison.} The evaluation spans three aspects: editing fidelity, user preference, and reconstruction fidelity. CLIP-based metrics~\cite{radford2021learning} show that Dynamic-eDiTor achieves strong alignment with the editing prompts across 4D scenes, and user studies indicate a clear preference for our results over the baselines in terms of semantic alignment, perceptual realism, and coherent motion. Although reconstruction metrics (PSNR, SSIM~\cite{wang2004image}, LPIPS~\cite{zhang2018unreasonable}) are slightly lower, they remain competitive and do not detract from the method’s overall superiority in semantic accuracy and perceptual edit quality.}
    \label{tab:comparison}
    \vspace{-1.5em}
\end{table*}

In this process, the token representation is defined as \(\phi(\mathcal{S}_{v,t}) = \mathrm{STGA}(\mathcal{S}_{v,t})\). We employ two Context Token Propagation strategies: Full Token Inheritance and Flow-guided Token Replacement. Full Token Inheritance is applied when the current sub-grid \(S_{curr}\) shares frames with the previous sub-grid \(S_{prev}\) along the temporal axis \((t = 1 \rightarrow T-1)\) or the spatial axis \((v = 1 \rightarrow V-1)\). We directly replace the entire current token \(\phi(\mathcal{S}_{curr})\) in these overlapped frames with previous token \(\phi(\mathcal{S}_{prev})\).

For a sub-grid along the temporal axis, a defined traversal path yields non-overlapped regions in the rightmost column of the sub-grid. Thus, we apply Flow-guided Token Replacement to these regions, in which the tokens are replaced with tokens warped from the corresponding rightmost column regions of the previous sub-grid. To ensure temporal alignment during warping, we estimate forward and backward optical flow between frames $f_t$ and $f_{t-1}$ using RAFT~\cite{teed2020raft}, and downsample the flow fields to match the token resolution. For each spatial location $(x,y)$, we use the downsampled forward flow $\mathbf{F}_{t \rightarrow t-1}(x,y)$ to backward-warp the tokens from the previous patch:
{
\setlength{\abovedisplayskip}{5pt}
\setlength{\belowdisplayskip}{5pt}
\setlength{\abovedisplayshortskip}{5pt}
\setlength{\belowdisplayshortskip}{5pt}
\begin{equation}
\hat{\phi}_{{\mathrm{r}}}(\mathcal{S}_{v,t}) \;=\;
\mathrm{Warp}\big(\mathbf{F}_{t \rightarrow t-1}(x,y),\, \phi_{{\mathrm{r}}}(\mathcal{S}_{v,t-1})\big).
\label{eq:token_warping}
\end{equation}
}
where $\hat{\phi}_{\mathrm{r}}(\mathcal{S}_{v,t})$ denotes the warped tokens in the rightmost column of the patch. During the replacement, we compute a validity mask \(\mathrm{M(x,y)}\) via a forward–backward consistency check, inspired by~\cite{meister2018unflow, xu2022gmflow}, to ensure precise replacement. With the mask \(\mathrm{M}\), tokens in valid regions are replaced by the warped tokens, 
while those in invalid regions retain the current frame’s tokens:
{
\setlength{\abovedisplayskip}{5pt}
\setlength{\belowdisplayskip}{5pt}
\setlength{\abovedisplayshortskip}{5pt}
\setlength{\belowdisplayshortskip}{5pt}
\begin{equation}
\phi_{{\mathrm{r}}}(\mathcal{S}_{v,t}) = \mathrm{M} \odot \hat{\phi}_{\mathrm{r}}(\mathcal{S}_{v,t}) + (1 - \mathrm{M}) \odot \phi_{\mathrm{r}}(\mathcal{S}_{v,t}),
\label{eq:replacement}
\end{equation}
}
where \(\odot\) denotes element-wise multiplication.

This unified propagation mechanism enables efficient and robust propagation across both spatial and temporal dimensions. STGA provides locally coherent spatial-temporal feature aggregation, while CTP propagates this coherence globally across the entire \(Grid\). As a whole, they ensure consistent motion and geometry in multi-view videos, significantly improving stability in 4D editing.

\subsection{Direct 4D Scene Optimization}
Our approach produces multi-view and temporally consistent edited video frames, which can be directly used to optimize the pre-trained 4D representation. 
In contrast to prior works~\cite{he2025ctrl, mou2024instruct} that rely on the Iterative Dataset Update (IDU), 
we directly optimize the 4D Gaussian model \(\mathcal{G}'_{\text{edit}}\) using all edited frames \(f_{v,t}^{\text{edit}}\) across the entire \(Grid\). The optimization objective is defined as:
{
\setlength{\abovedisplayskip}{5pt}
\setlength{\belowdisplayskip}{5pt}
\setlength{\abovedisplayshortskip}{5pt}
\setlength{\belowdisplayshortskip}{5pt}
\begin{equation}
\mathcal{G}'_{\text{edit}} =
\arg\min_{\mathcal{G}}
\sum_{v,t \in V,T}
\left\lVert \hat{f}_{v,t} - f_{v,t}^{\text{edit}} \right\rVert
+ \mathcal{L}_{\text{tv}},
\label{eq:direct_opt}
\end{equation}
}
where both loss terms follow the 4DGS~\cite{wu20244d}. 


\begin{table*}[htbp]
\footnotesize
    \centering
    \renewcommand\arraystretch{1.0}
    \setlength{\tabcolsep}{3.1mm}{
    \begin{threeparttable}
    \begin{tabular}{ccccccccc}
        \toprule
        & & \multicolumn{2}{c}{\textbf{2D Consistency}} 
        & \multicolumn{3}{c}{\textbf{Reconstruction Fidelity}} 
        & \multicolumn{2}{c}{\textbf{Editing Fidelity}} \\
        \cmidrule(lr){3-4} \cmidrule(lr){5-7} \cmidrule(lr){8-9}
        \textbf{STGA} &
        \textbf{CTP} &
        \makecell{Warp-Err $\scriptscriptstyle 10^{-3}$} $\downarrow$ &
        \makecell{MEt3R $\scriptstyle 10^{-1}$}  $\downarrow$ &
        \makecell{PSNR} $\uparrow$ &
        \makecell{SSIM} $\uparrow$ &
        \makecell{LPIPS} $\downarrow$ &
        \makecell{CLIP$_{\text{dir}}$} $\uparrow$ &
        \makecell{CLIP$_{\text{sim}}$} $\uparrow$ \\
        \cmidrule(lr){1-2} \cmidrule(lr){3-4} \cmidrule(lr){5-7} \cmidrule(lr){8-9}
         - & - & 56.98 & 1.0721 & 26.14 & 0.7445 & 0.1408 & \underline{0.1930} & {0.5414} \\
        \checkmark & - & 38.64 & \underline{0.9277} & 28.08 & 0.7875 & \underline{0.1122} & {0.1872} & \underline{0.6407} \\
         - & \checkmark & \underline{29.44} & 1.0695 & \underline{28.74} & \underline{0.8013} & 0.1165 & \textbf{0.1944} & \textbf{0.6418}\\
        \cmidrule(lr){1-2} \cmidrule(lr){3-4} \cmidrule(lr){5-7} \cmidrule(lr){8-9}
        \checkmark & \checkmark &  \textbf{28.94} & \textbf{0.9074} & \textbf{29.25} & \textbf{0.8064} & \textbf{0.1006}  & 0.1849 & 0.6397 \\
        \bottomrule
    \end{tabular}
    \end{threeparttable}}
    \vspace{-0.5em}
    \caption{\textbf{Ablation Study.} Each component, Spatio-Temporal Sub-Grid Attention (STGA) and Context Token Propagation (CTP), helps preserve temporal and multi-view consistency, improving the 4D reconstruction quality. Our method prioritizes a globally stable 4D structure, yielding consistent temporal and spatial behavior and thus more robust reconstruction fidelity. Although CLIP-based metrics~\cite{radford2021learning} show a slight drop due to the trade-off between semantic alignment and spatio-temporal coherence, our method still produces more stable and reliable 4D edits, avoiding the geometric and temporal artifacts seen in the ablated variants.}
    \label{tab:ablation}
\end{table*}

\vspace{-1.0em}
\begin{table*}[htbp]
\footnotesize
    \centering
    \renewcommand\arraystretch{1.0}
    \setlength{\tabcolsep}{2.5mm}{
    \begin{threeparttable}
    \begin{tabular}{ccccccccc}
        \toprule
        & & \multicolumn{2}{c}{\textbf{2D Consistency}} 
        & \multicolumn{3}{c}{\textbf{Reconstruction Fidelity}} 
        & \multicolumn{2}{c}{\textbf{Editing Fidelity}} \\
        \cmidrule(lr){3-4} \cmidrule(lr){5-7} \cmidrule(lr){8-9}
        \makecell{\textbf{CTP-Full}} &
        \makecell{\textbf{CTP-Flow}} &
        \makecell{Warp-Err $\scriptscriptstyle 10^{-3}$}  $\downarrow$&
        \makecell{MEt3R $\scriptscriptstyle 10^{-1}$ }  $\downarrow$ &
        \makecell{PSNR $\uparrow$} &
        \makecell{SSIM $\uparrow$} &
        \makecell{LPIPS $\downarrow$} &
        \makecell{CLIP$_\text{dir}$ $\uparrow$} &
        \makecell{CLIP$_\text{sim}$ $\uparrow$} \\
        \cmidrule(lr){1-2} \cmidrule(lr){3-4} \cmidrule(lr){5-7} \cmidrule(lr){8-9}
    
        - & - & 38.64 & 0.9277 & 28.08 & 0.7875 & 0.1122 & \textbf{0.1872} & \textbf{0.6407} \\
        - & \checkmark & \underline{29.79} & 0.9205 & \underline{28.97} & \underline{0.7990} & \underline{0.1034} & 0.1852 & \underline{0.6402} \\
        \checkmark & - & 33.22 & \underline{0.9094} & 28.19 & 0.7906 & 0.1089 & \underline{0.1865} & 0.6400 \\
        \cmidrule(lr){1-2} \cmidrule(lr){3-4} \cmidrule(lr){5-7} \cmidrule(lr){8-9}
        \checkmark & \checkmark & \textbf{28.94} & \textbf{0.9074} & \textbf{29.25} & \textbf{0.8064} & \textbf{0.1006} & 0.1849 & 0.6397 \\
        \bottomrule
    \end{tabular}
    \end{threeparttable}}
    \vspace{-0.5em}
    \caption{\textbf{Ablation Study: Context Token Propagation (CTP).} This ablation study is conducted with STGA included to isolate the impact of CTP. Full Token Inheritance (CTP-Full) and Flow-Guided Token Replacement (CTP-Flow) play a critical role in reinforcing temporal and multi-view consistency, enabling more accurate reconstruction of the edited dynamic scene. Despite a slight trade-off in CLIP-based metrics~\cite{radford2021learning}, CTP substantially improves spatio-temporal coherence and overall 4D editing fidelity.}
    \vspace{-2.em}
    \label{tab:ablation_token}
\end{table*}

\vspace{0.5em}
\section{Experiment}
\subsection{Experimental Setup}
\paragraph{Dataset.}

We evaluate our method on the real-world multi-view video dataset DyNeRF~\cite{li2022neural}, which contains six dynamic scenes with 16-21 camera views capturing 10-second videos at 30 FPS.
To evaluate editing consistency under sparse temporal sampling, we uniformly sample frames at 1FPS (160-210 frames per scene, compared to 4,800-6,300 frames at 30FPS). All experiments are conducted using 14 prompts covering all six scenes in the dataset. 



\vspace{-1.5em}
\paragraph{Baselines.}
We compare our method against state-of-the-art 4D scene editing approaches, including Instruct 4D-to-4D~\cite{mou2024instruct}, CTRL-D~\cite{he2025ctrl}, and Instruct 4DGS~\cite{kwon2025efficient}. Since our task focuses on text-based scene editing, we reproduce all baseline results using text prompt input only. 


\begin{figure}[t]
    \centering
    \includegraphics[width=\linewidth]{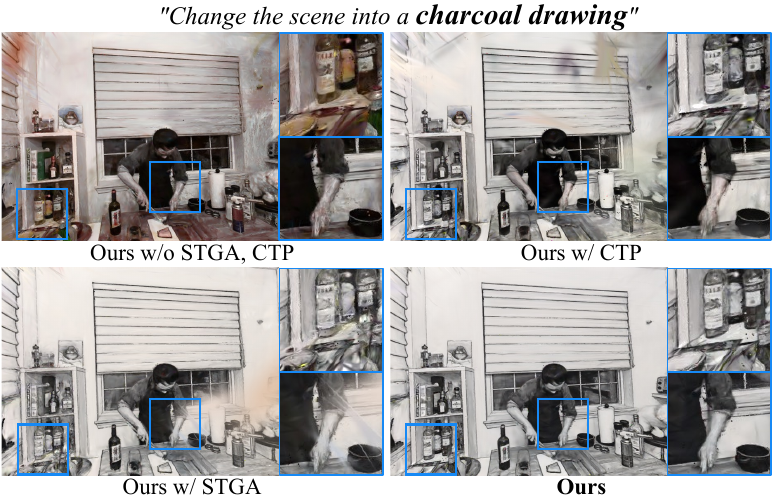}
    \vspace{-2.em}
    \caption{
    \textbf{Qualitative Ablation Results.} The model lacking both components (top-left) suffers from severe artifacts and geometric drift. Adding only STGA or only CTP progressively improves the result, but still leaves residual motion blur and geometric drift. Our full method (bottom-right) successfully ensuring the spatio-temporal consistency to produce a stable and complete edit.}
    \label{fig:ablation1}
    \vspace{-1.7em}
\end{figure}

\subsection{Implementation Details}
Our method leverages the MM-DiT-based~\cite{esser2024scaling} image editing model, Qwen-Image-Edit~\cite{wu2025qwen} and 4D Gaussian Splatting~\cite{wu20244d}. For evaluation, we use all camera views in the dataset and hold out the final frame of each view as the test set. We evaluate our rendered results on this test set. The full 4D editing process takes approximately 51 minutes for the ``coffee martini" scene on a single NVIDIA H100 GPU.

\subsection{Qualitative Results}
~\cref{fig:teaser} illustrates Dynamic-eDiTor's ability to perform multi-view temporal scene editing. Our model effectively edits diverse scenes and local objects while maintaining strong temporal and spatial consistency. Dynamic-eDiTor is able to perform non-rigid appearance editing, semantic local editing, and consistent stylization while still preserving motion consistency across viewpoints and over time. 

In ~\cref{fig:comparison}, we compare Dynamic-eDiTor with recent 4D scene editing baselines~\cite{mou2024instruct, kwon2025efficient, he2025ctrl}. We observe that most baseline methods fail to handle non-rigid content manipulation and are often limited to style-oriented edits. This core limitation leads to significant artifacts, such as motion distortions, geometric drift, and incomplete editing results. These failures are evident across the examples. When editing the scene into a “fire emergency”, all baselines fail to generate plausible emergency-related elements, revealing weak text–scene alignment and incomplete editing. In the second column, Instruct-4DGS struggles with non-rigid editing, causing clear motion distortions around the hand. Meanwhile, Instruct 4D-to-4D and CTRL-D introduce noticeable artifacts such as facial color shifts and blurring. In the third column, CTRL-D further demonstrates viewpoint inconsistencies and produces blurred edited regions, while other baselines result in incomplete edits. Instruct 4D-to-4D fails to modify the target scene, incorrectly altering surrounding objects. This indicates weak text alignment despite sharing the same diffusion backbone such as InstructPix2Pix~\cite{brooks2023instructpix2pix} as other baselines. Overall, Dynamic-eDiTor outperforms all previous 4D scene editing methods by achieving superior editing completeness and effectively preserving temporal coherence and multi-view consistency, resulting in high-quality dynamic scene edits.

\subsection{Quantitative Results}
~\cref{tab:comparison} presents a quantitative comparison with prior 4D scene editing methods, focusing on the 4D rendered image quality. Our evaluation is structured into two categories: text-prompt alignment and reconstruction fidelity.

To evaluate text-prompt alignment, we use CLIP-based ~\cite{radford2021learning} metrics. Specifically, the CLIP text-image directional similarity captures how changes in text captions correspond to changes between the source and rendered images in CLIP embedding space, while the CLIP text-image similarity directly measures alignment between the target text and rendered frames. Our method surpasses all baselines in these CLIP metrics, demonstrating that our rendered results are significantly better aligned with the user's intended edit.

To assess reconstruction fidelity, we report PSNR, SSIM~\cite{wang2004image}, and LPIPS~\cite{zhang2018unreasonable}. These metrics are computed between the final rendered test frames and the corresponding 2D edited target frames, measuring how faithfully the 4D model reconstructs the target edits. Although Dynamic-eDiTor obtains slightly lower values than CTRL-D on these reconstruction metrics, this highlights that our method achieves a better balance by prioritizing faithful text alignment and reliable 4D scene editing. This trade-off is further supported by our vital layer analysis in ~\cref{fig:layer_analysis}.

Beyond reconstruction metrics, we evaluate perceptual quality through a user study with 150 participants on Amazon Mechanical Turk~\cite{amazonturk}. Participants were asked to compare our final 4D rendered videos against baseline methods~\cite{mou2024instruct, he2025ctrl, kwon2025efficient}. As shown in~\cref{tab:comparison}, our method consistently outperforms the baselines in terms of overall visual quality,  motion consistency, temporal and multi-view consistency, text-prompt alignment, and identity preservation.

\begin{figure}[t]
    \centering
    \includegraphics[width=\linewidth]{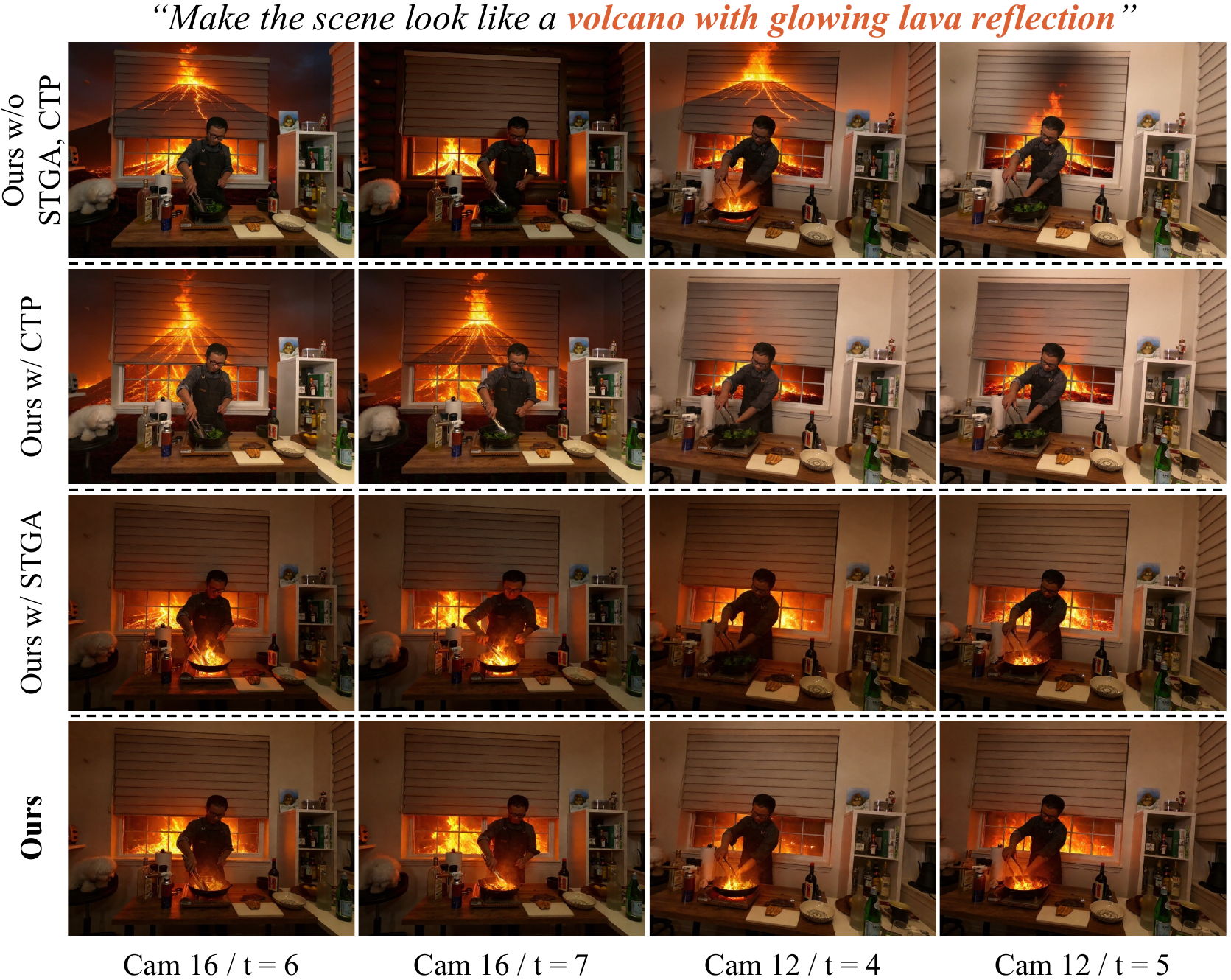}
    \vspace{-2.em}
    \caption{\textbf{Ablation Study: 2D Consistency.} Each component in our method strengthens temporal and multi-view consistency in 2D editing. STGA improves semantic alignment and preserves fine details across views, while CTP enhances coherence by propagating information across neighboring frames.} 
    \label{fig:ablation_2d}
    \vspace{-2.0em}
\end{figure}


\subsection{Ablation Study}
We conduct an ablation study on our Grid-based Spatio-Temporal Propagation and Context Token Propagation.

\noindent\textbf{2D Consistency.} We first assess each component’s impact on 2D temporal and multi-view consistency, a key factor for high-quality 4D reconstruction. As shown in ~\cref{fig:ablation_2d}, STGA strengthens semantic alignment and view consistency, whereas CTP improves temporal coherence through information propagation. Collectively, they yield notable improvements in 2D spatial and temporal stability. The quantitative results in ~\cref{tab:ablation} support these findings. Our full method achieves superior spatio-temporal consistency, evidenced by the lowest warping error~\cite{lai2018learning} and MEt3R~\cite{asim2025met3r}, which further strengthens overall 4D fidelity.

\noindent\textbf{4D Fidelity.} ~\cref{fig:ablation1} demonstrates how improved 2D consistency translates into higher-quality 4D scene edits. Without our components, the edited scene exhibits severe motion artifacts, especially around the man's hand, along with background degradation and incorrect text alignment, such as failing to produce “charcoal drawing” colors. Adding STGA reduces large-scale artifacts and stabilizes dynamic motion, while incorporating CTP further refines fine-grained details by leveraging temporal and multi-view cues from previous grids. With all components combined, Dynamic-eDiTor achieves consistent motion reconstruction and editing, effectively eliminating geometric drift in the 4D rendered output. ~\cref{tab:ablation} reveals that each component reinforces 4D fidelity. The PSNR, SSIM, and LPIPS scores validate this, demonstrating that Dynamic-eDiTor delivers highly coherent edits. We also note that CLIP-based metrics are slightly higher when STGA is removed. As mentioned in \cref{vital_layer}, this reflects the trade-off between semantic alignment and spatio-temporal coherence. balanced, spatio–temporally coherent 4D reconstruction, rather than optimizing semantic alignment alone. Our method prioritizes a balanced, stable, and coherent 4D reconstruction, whereas the ablated variant attains higher CLIP-based metrics by sacrificing this stability, resulting in geometrically and temporally unstable reconstrution.

\begin{figure}[t]
    \centering
    \includegraphics[width=\linewidth]{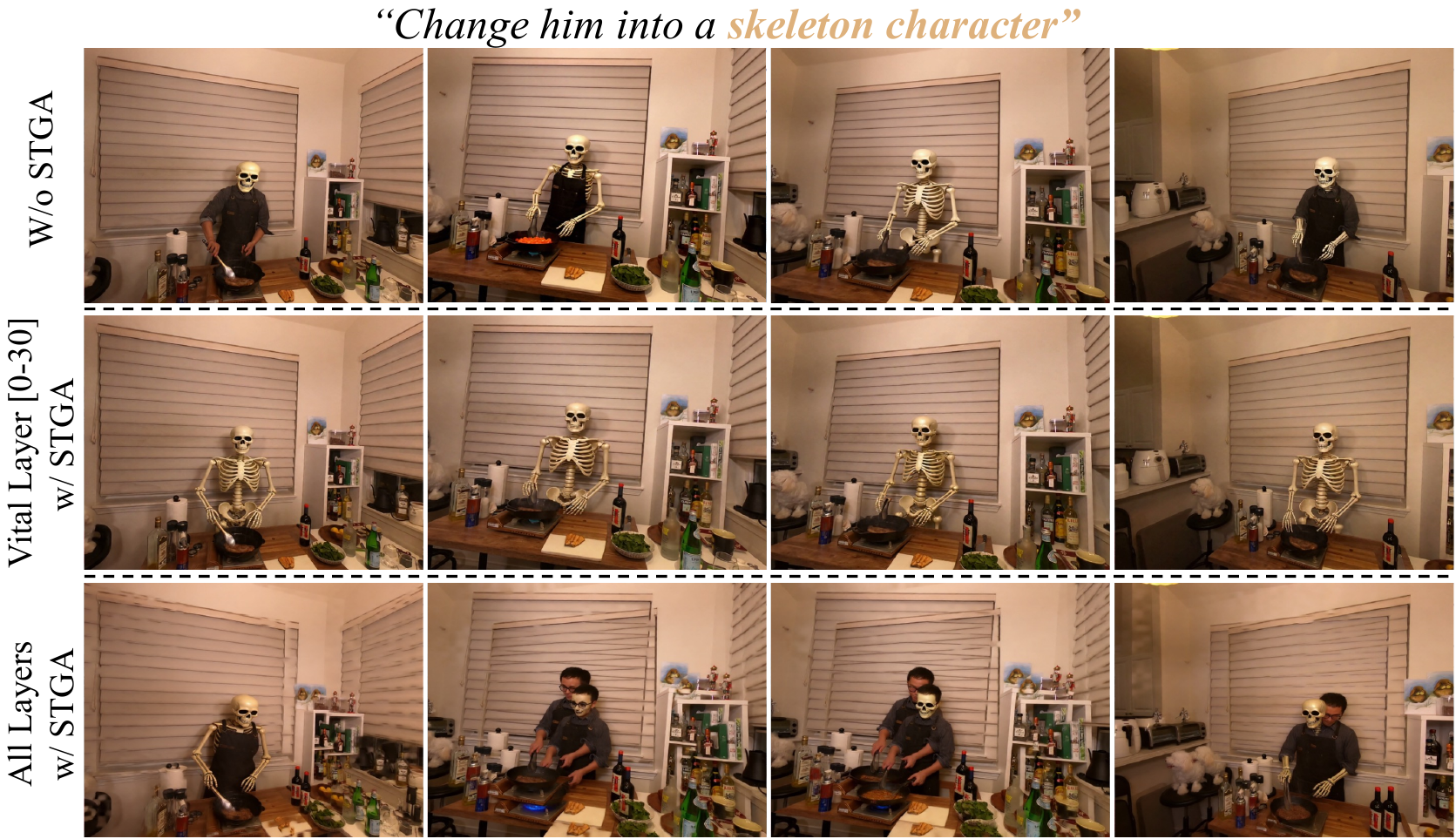}
    \vspace{-2.em}
    \caption{\textbf{Qualitative Analysis of Vital Layer Range.} Applying STGA to all layers introduces visual artifacts across views and time, while omitting STGA produces inconsistent multi-view and temporal edits. Restricting STGA to the vital range yields the most coherent and stable multi-view–time editing results.}
    \label{fig:ablation_vital}
    \vspace{-1.5em}
\end{figure}

\noindent\textbf{Context Token Propagation.}
We further ablate our Context Token Propagation (CTP) components in ~\cref{tab:ablation_token}. First, removing only the Full Token Inheritance leads to reduced 2D multi-view consistency and lower fidelity in the 4D rendered images. Next, removing only the Flow-guided Token Replacement results in a significant drop in temporal consistency. Finally, removing the entire Context Token Propagation mechanism (both components) causes a severe degradation in performance, dramatically worsening both 2D consistency and 4D reconstruction fidelity. Similar to the previous ablation, these results reflect the inherent trade-off between semantic alignment and spatio-temporal coherence, confirming that both the Full Token Inheritance and Flow-guided Token Replacement are essential for achieving high-quality and consistent 4D editing.





\vspace{0.5em}
\section{Conclusion}
We presented Dynamic-eDiTor, a training-free framework for text-driven 4D scene editing that achieves spatially and temporally consistent results across multi-view videos, enabling stable optimization of 4D representations with MM-DiT~\cite{esser2024scaling, wu2025qwen} and 4DGS~\cite{wu20244d}. The core of our approach is Grid-based Spatio-Temporal Propagation, combining Spatio-Temporal Sub-Grid Attention (STGA) for localized view-time fusion and Context Token Propagation (CTP) for explicit global consistency. Together, these components ensure coherent geometry and motion, and high-fidelity dynamic edits. Extensive experiments on DyNeRF~\cite{li2022neural} demonstrate our method significantly outperforms prior work in editing fidelity, temporal smoothness, and robustness. We believe Dynamic-eDiTor represents a notable progression toward text-driven dynamic scene editing. 

\noindent \textbf{Limitation.}
While Dynamic-eDiTor effectively enforces multi-view and temporal consistency during text-driven edits, it is less suitable for large-scale geometric alterations such as substantial motion reconfiguration or topology-changing edits. Since our framework operates by propagating spatio-temporal features without modeling geometric deformation, edits that require significant structural changes remain challenging. Extending our approach to handle more drastic motion editing or geometry-changing transformations represents an important direction for future work.

\section{Acknowledgement}
{
This work is partially supported by the NSF under the Future of Work at the Human-Technology Frontier (FW-HTF) 1839971 and Partnership for Innovation (NSF PFI 2329804). The authors also acknowledge the Feddersen Distinguished Professorship Funds. Additional support for this work is partially funded by the Culture, Sports, and Tourism R\&D Program through the Korea Creative Content Agency grant funded by the Ministry of Culture, Sports and Tourism in 2024~(International Collaborative Research and Global Talent Development for the Development of Copyright Management and Protection Technologies for Generative AI, RS-2024-00345025, 25\%),
by the Korea government(MSIT) the National Research Foundation of Korea(NRF) grant funded (RS-2025-00521602), the Institute of Information \& Communications Technology Planning \& Evaluation (IITP)(RS-2019-II190079, 1\%), and
Artificial intelligence industrial convergence cluster development project funded by the Ministry of Science and ICT(MSIT, Korea) \& Gwangju Metropolitan City. Any opinions, findings, and conclusions expressed in this material are those of the authors and do not necessarily reflect the views of the funding agency.}


{
    \small
    \bibliographystyle{ieeenat_fullname}
    \bibliography{main}
}

\clearpage
\setcounter{page}{1}
\maketitlesupplementary
\setcounter{section}{0}
\renewcommand{\thesection}{\Alph{section}}

\section*{Overview}
This supplementary material provides additional details, analyses, and experimental results for our proposed method, Dynamic-eDiTor.
\vspace{0.5em}

\begin{itemize}

\item \cref{fig:supp_comparison} and \cref{fig:supp_qualitative_result} present additional qualitative results, including extended comparisons with baseline methods.
\vspace{0.5em}

\item \cref{supp:imple} and \cref{supp:metric} provide further implementation details and descriptions of all evaluation metrics used in our experiments.
\vspace{0.5em}

\item \cref{supp:userstudy} summarizes the user study protocol and provides a detailed analysis of participant responses.
\vspace{0.5em}

\item \cref{supp:traversal} details our Grid-based Spatio-Temporal Propagation mechanism, including Asymmetric Traversal Strategy and the accompanying algorithm in \cref{alg:subgrid_traversal}.
\vspace{0.5em}

\item \cref{supp:vital} offers an extended analysis of the \emph{vital layer range} for Spatio-Temporal Sub-Grid Attention (STGA).
\vspace{0.5em}

\item \cref{supp:ablation} contains additional ablation studies analyzing Asymmetric Traversal Strategy of Dynamic-eDiTor.
\vspace{0.5em}

\item \cref{supp:mono} presents additional qualitative results in the monocular video setting of 4D Gaussian Splatting (4DGS)~\cite{wu20244d}.
\vspace{0.5em}

\end{itemize}

\section{Implementation Details}
\label{supp:imple}
For each scene, we first train the source 4D Gaussian Splatting~\cite{wu20244d} representation for 30{,}000 iterations using the Adam optimizer~\cite{kingma2014adam} with the same learning rate schedule as 4DGS. During the editing stage, we optimize the model for 20{,}000 iterations using the edited frames, following the original 4DGS hyperparameter configuration. All experiments are conducted on an NVIDIA H100 GPU; however, by employing local caching for Temporal Context Token Replacement, our method also runs efficiently on an NVIDIA A6000 GPU.

For the 2D MM-DiT~\cite{esser2024scaling,wu2025qwen} image editor, we utilize Qwen-Image-Edit~\cite{wu2025qwen} from the Diffusers library~\cite{von-platen-etal-2022-diffusers}. To enhance computational efficiency, we incorporate the LoRA~\cite{hu2022lora} weight Qwen-Image-Lightning-8steps-V1.1. All input images are resized to 768 × 768 before processing.

\begin{algorithm}[t]
\caption{Asymmetric Sub-Grid Traversal}
\label{alg:subgrid_traversal}
\begin{algorithmic}[1]
\Require Camera--time grid $\mathrm{Grid} = \{ f_{v,t} \}$ of size $V \times T$
\Ensure Ordered list of sub-grids $\Omega = \{ \mathcal{S}^{(k)} \}$
\State $\Omega \gets [\ ]$ \Comment{Initialize empty sub-grid sequence}
\For{$v = 0$ \textbf{to} $V - 2$}
    \If{$v$ is even \textbf{or} $v = V - 2$}
        \Comment{Temporal sweep}
        \For{$t = 0$ \textbf{to} $T - 2$}
            \State $\mathcal{S}_{v,t} \gets \{ f_{v,t}, f_{v+1,t}, f_{v,t+1}, f_{v+1,t+1} \}$
            \State Append $\mathcal{S}_{v,t}$ to $\Omega$
        \EndFor
    \Else
        \Comment{Cross-view alignment at $t=0$}
        \State $\mathcal{S}_{v,0} \gets \{ f_{v,0}, f_{v+1,0}, f_{v,1}, f_{v+1,1} \}$
        \State Append $\mathcal{S}_{v,0}$ to $\Omega$
    \EndIf
\EndFor
\State \textbf{return} $\Omega$
\end{algorithmic}
\end{algorithm}

\begin{figure*}[t]
    \vspace{-2.2em}
    \centering
    \includegraphics[width=\linewidth]{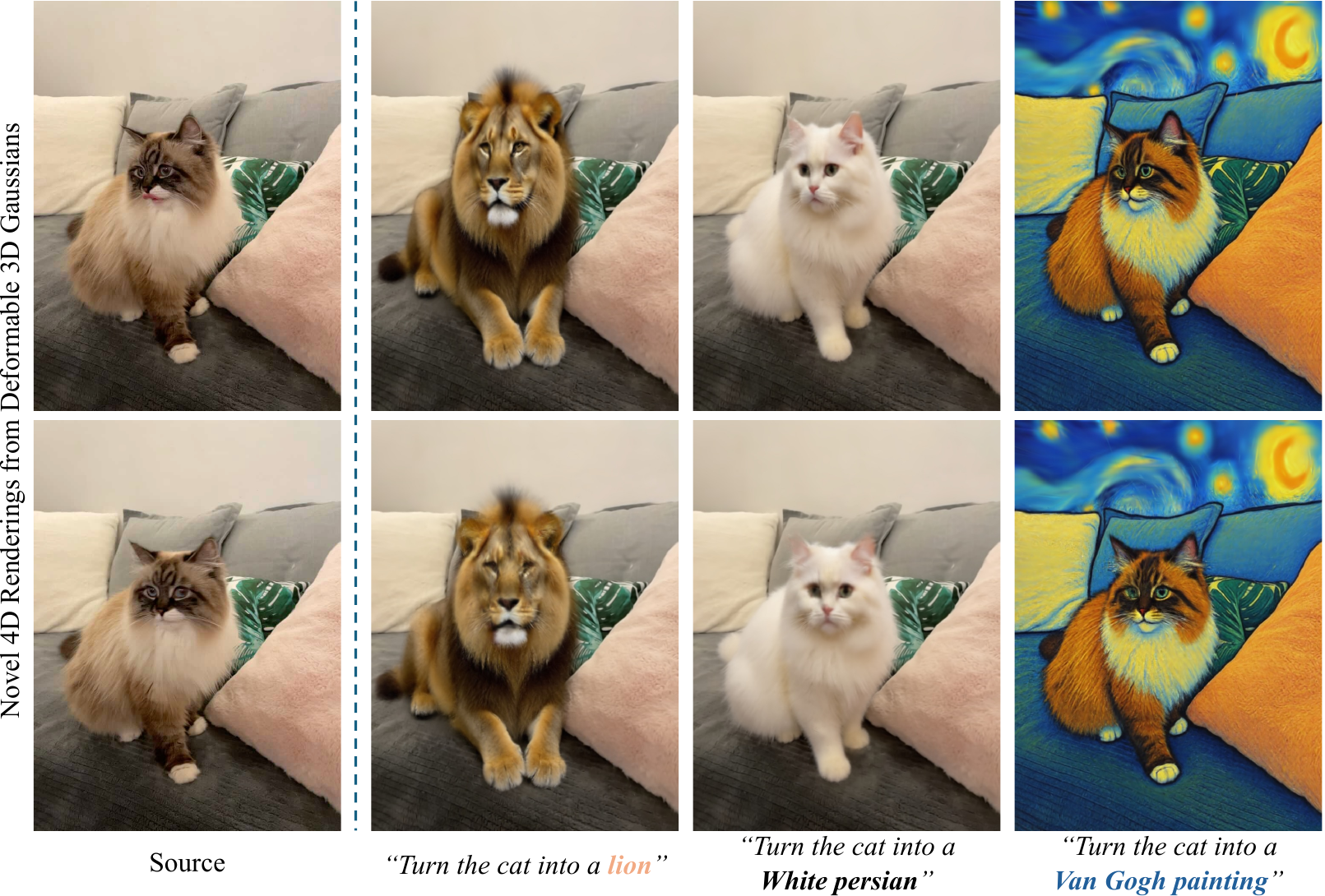}
    \vspace{-2.em}
    \caption{\textbf{Qualitative Results in Monocular video setting.} We evaluate the applicability of Dynamic-eDiTor in the challenging monocular video setting, where only a single moving-camera sequence is available and no multi-view redundancy exists. Using Deformable 3D Gaussian Splatting~\cite{yang2023deformable3dgs} as the underlying 4D representation, our method successfully performs text-driven appearance and object manipulations while maintaining stable geometry and consistent motion over time. } 
    \vspace{-1.em}
    \label{fig:monocular}
\end{figure*}

For baseline comparisons, we follow the official implementations of Instruct4D-to-4D~\cite{mou2024instruct} and Instruct-4DGS~\cite{kwon2025efficient}, since both are text-driven 4D editing methods comparable to ours.
CTRL-D~\cite{he2025ctrl} requires an additional edited reference image that must be produced by choosing one of several diffusion-based editing modes, such as image-prompt editing, text-prompt editing, or mask-based editing, before fine-tuning its InstructPix2Pix~\cite{brooks2023instructpix2pix} backbone.
To ensure a fair and consistent comparison, we fix this pre-editing stage to use the standard InstructPix2Pix image editor, which is the backbone originally used in CTRL-D, when generating the reference edited image.

\section{Metric}
\label{supp:metric}
To evaluate Dynamic-eDiTor, we use a combination of 2D consistency, 4D editing fidelity, and 4D reconstruction fidelity metrics.

For \textbf{2D consistency}, we evaluate both temporal and multi-view stability. 
Our 4D editing baselines such as Instruct4D-to-4D~\cite{mou2024instruct} and CTRL-D~\cite{he2025ctrl} rely on Iterative Dataset Update (IDU)\cite{haque2023instruct}, and Instruct-4DGS\cite{kwon2025efficient} uses an SDS-based~\cite{poole2022dreamfusion} optimization strategy. However, these approaches do not generate temporally aligned or viewpoint-consistent 2D edited frames. Their updates are stochastic and occur directly in 3D or 4D space, which makes extracting coherent multi-view video sequences infeasible. Therefore, 2D consistency metrics cannot be fairly compared with these baselines and are used only within our ablation studies.

\begin{itemize}
    \item \textbf{MEt3R}~\cite{asim2025met3r}:  
        Evaluates multi-view consistency by comparing feature similarity between view-warped images.  
        We employ the official MEt3R metric with MASt3R~\cite{leroy2024grounding}, DINOv2~\cite{oquab2023dinov2} (FeatUp) features, 448 image resolution, and cosine similarity.  
        Lower values indicate more coherent appearance across viewpoints.

    \item \textbf{Warping Error}~\cite{lai2018learning}:  
    Measures temporal consistency by computing the discrepancy between frame $f_{t}$ and the optical-flow–warped version of frame $f_{t-1}$ using RAFT~\cite{teed2020raft}.  
    Lower scores indicate smoother temporal alignment and fewer motion artifacts.
    
\end{itemize}

\begin{table*}[htbp]
\small
    \centering
    \renewcommand\arraystretch{1.0}
    \setlength{\tabcolsep}{1.5mm}{
    \begin{threeparttable}
    \begin{tabular}{cccccccccccc}
        \toprule
        & & \multicolumn{4}{c}{\textbf{2D Consistency}} 
        & \multicolumn{3}{c}{\textbf{Reconstruction Fidelity}} 
        & \multicolumn{2}{c}{\textbf{Editing Fidelity}} \\
        \cmidrule(r){3-6} \cmidrule(r){7-9} \cmidrule(r){10-11}
        \textbf{\footnotesize AGT} &
        \textbf{\footnotesize STGA} &
        \makecell{\footnotesize Local\\Warp-Err $\scriptscriptstyle 10^{-3}$} $\downarrow$ &
        \makecell{\footnotesize Global\\Warp-Err $\scriptscriptstyle 10^{-3}$} $\downarrow$ &
        \makecell{\footnotesize Local\\MEt3R $\scriptstyle 10^{-1}$}  $\downarrow$ &
        \makecell{\footnotesize Global\\MEt3R $\scriptstyle 10^{-1}$} $\downarrow$ &
        \makecell{\footnotesize PSNR} $\uparrow$ &
        \makecell{\footnotesize SSIM} $\uparrow$ &
        \makecell{\footnotesize LPIPS} $\downarrow$ &
        \makecell{\footnotesize CLIP$_{\text{dir}}$} $\uparrow$ &
        \makecell{\footnotesize CLIP$_{\text{sim}}$} $\uparrow$ \\
        \cmidrule(r){1-2} \cmidrule(r){3-6} \cmidrule(r){7-9} \cmidrule(r){10-11}
        - & - & 56.98 & 58.47 & 1.0721 & 1.4312 & 26.14 & 0.7445 & 0.1408 & \textbf{0.1930} & \underline{0.6414} \\
         - & \checkmark & \textbf{34.56} & \underline{42.86} & \textbf{0.9266} & \underline{1.2984} & \underline{27.84} & \underline{0.7793} & \underline{0.1160} & \underline{0.1876} & \textbf{0.6499} \\
        \checkmark & \checkmark & \underline{38.64} & \textbf{42.33} & \underline{0.9277} & \textbf{1.2953} & \textbf{28.08} & \textbf{0.7875} & \textbf{0.1122} & {0.1872} & {0.6407} \\
        \bottomrule
    \end{tabular}
    \end{threeparttable}}
    \caption{\textbf{Ablation Study: Asymmetric Sub-Grid Traversal (AGT).} This evaluation is conducted without CTP to isolate the impact of Asymmetric Sub-Grid Traversal (AGT). The results show that sub-grids without AGT achieve slightly better local consistency metrics because all frames within each sub-grid are updated independently. However, the lack of linkage between sub-grids introduces discontinuities, weakening overall 4D reconstruction fidelity. In contrast, applying AGT improves global consistency by overlapping frames across sub-grids, even at the cost of some local editing precision, as it enables effective information propagation. This leads to more stable and reliable 4D edits, demonstrating that global consistency is ultimately more critical for 4D reconstruction fidelity.}
    \vspace{-1.em}
    \label{tab:ablation_agt}
\end{table*}

For \textbf{4D editing fidelity}, we adopt CLIP-based metrics~\cite{radford2021learning}. We compute both CLIP text-image directional similarity and CLIP text-image similarity using the \emph{rendered images} produced by the edited 4D scene. The directional similarity evaluates whether the change described by the text prompt corresponds to the transformation from the source image to the edited rendering in CLIP embedding space. The CLIP text-image similarity, on the other hand, directly measures how well the rendered frames semantically align with the target text prompt.

For \textbf{4D reconstruction fidelity}, we report PSNR, SSIM~\cite{wang2004image}, and LPIPS~\cite{zhang2018unreasonable}, following prior works~\cite{doh2025occlusion, kwon2025efficient, zuo2024high, kirschstein2025avat3r}. All three metrics are computed between the edited test-view image and the rendered test-view image from the same camera viewpoint, enabling a direct comparison of reconstruction quality.

\section{User Study Detail}
\label{supp:userstudy}

To compare the editing performance of Dynamic-eDiTor against baseline methods, we conducted a user study with 150 participants on Amazon Mechanical Turk~\cite{amazonturk}. Each participant evaluated 14 scenarios, and for each scenario, they compared the 4D rendered video results produced by four systems across six subjective dimensions, as illustrated in~\cref{fig:supp_userstudy_QA}. We designed six evaluation questions covering prompt alignment (Q1), temporal consistency (Q2), viewpoint consistency (Q3), motion consistency (Q4), identity preservation (Q5), and overall visual quality (Q6). For each dimension, participants selected the system they judged to perform best. To reduce human bias, the presentation order of the four systems was randomized for every question.

For analysis, we first counted how many times Dynamic-eDiTor was selected across the 14 scenarios and compared it with the best-performing baseline (best baseline) on each dimension. The results show that Dynamic-eDiTor consistently outperformed the best baseline across all six evaluation dimensions. For example, on overall quality (Q6), Dynamic-eDiTor achieved an average selection rate of 0.49, compared to 0.28 for the best baseline (Instruct4d~\cite{mou2024instruct}). The advantage is even more pronounced for prompt alignment (Q1), with selection rates of 0.57 vs. 0.22 (Instruct4d). For other questions, Dynamic-eDiTor’s average selection rate exceeded the best baseline by approximately 0.17–0.21, demonstrating stable and comprehensive improvements.

To examine whether these differences were statistically significant, we performed a two-sided signed~\cite{dixon1946statistical} test for each question, pairing each participant’s selection ratio for Dynamic-eDiTor with that of the best baseline. All six dimensions yielded p-values far below 0.01, specifically: Q1 ($p = 8.88\times10^{-16}$), Q2 ($p = 5.43\times10^{-3}$), Q3 ($p = 4.49\times10^{-5}$), Q4 ($p = 1.41\times10^{-4}$), Q5 ($p = 9.33\times10^{-5}$), and Q6 ($p = 2.10\times10^{-5}$).

These results confirm that human evaluators consistently prefer Dynamic-eDiTor over the best baseline. The static significance also reveals that the gains are robust rather than due to random variation.

\section{Asymmetric Sub-Grid Traversal with Overlapping Structure}
\label{supp:traversal}

In the main paper, Grid-based Spatio-Temporal Propagation is introduced as a mechanism that performs local fusion via Spatio-Temporal Sub-Grid Attention (STGA) and global propagation via Context Token Propagation (CTP). 
In this section, we provide additional details on how the camera–time grid \(Grid\) is traversed and how overlapping sub-grids are constructed to enable stable spatio-temporal propagation. \cref{alg:subgrid_traversal} formalizes the sub-grid generation process used in our implementation.

\subsection{Overlapping Sub-Grids}

Since neighboring sub-grids share frames on their boundaries, they form an overlapping tiling of the camera--time grid.
This overlap is crucial for CTP: the shared regions act as “anchors” through which coherent token representations can be propagated from one sub-grid to the next.

\begin{figure}[t]
    \centering
    \includegraphics[width=\linewidth]{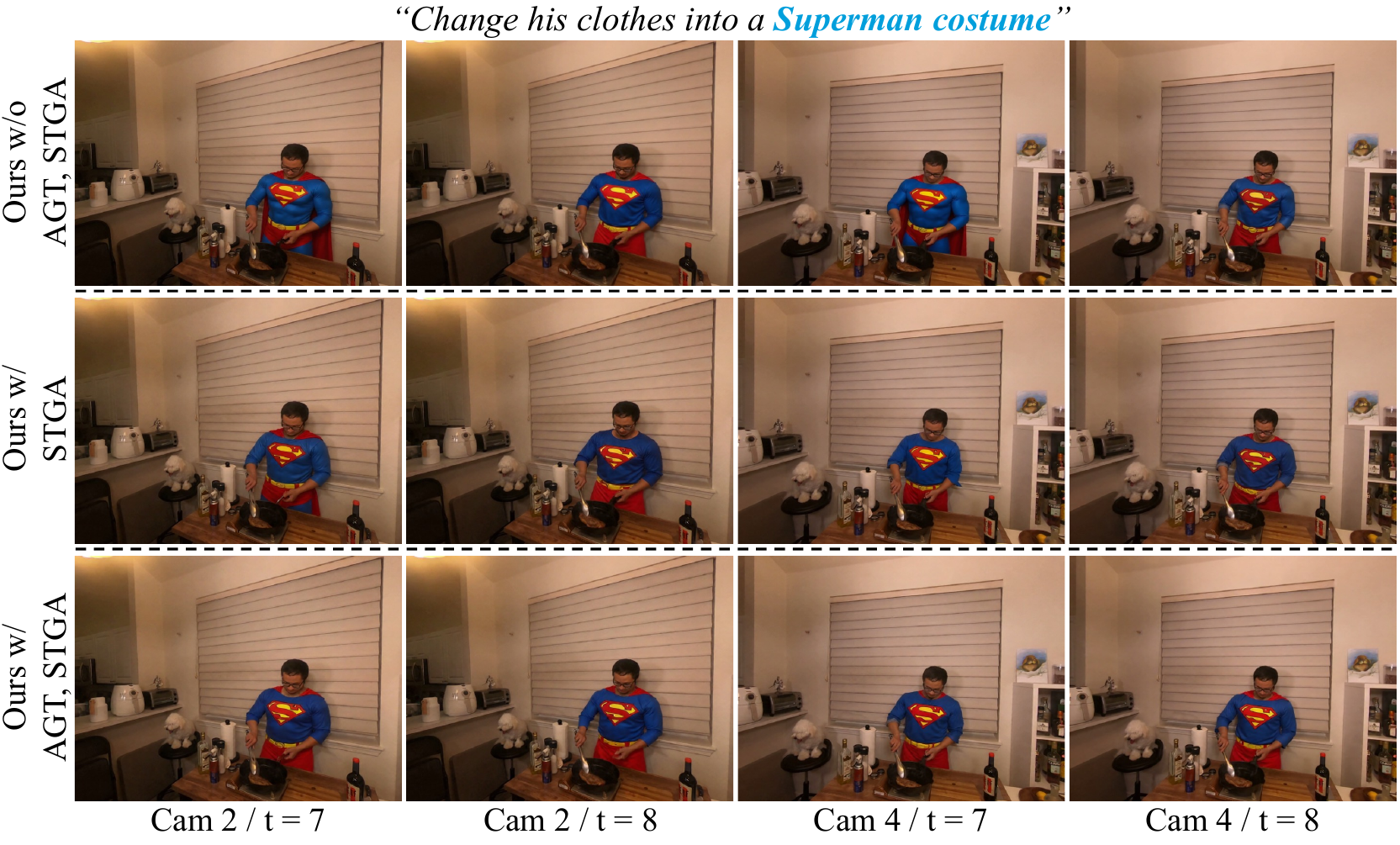}
    \vspace{-2.em}
    \caption{
    \textbf{Ablation Study: Asymmetric Sub-Grid Traversal (AGT).} This qualitative result clearly demonstrates that AGT preserves global multi-view and temporal consistency. Without AGT, noticeable discontinuities appear between sub-grids.} 
    \vspace{-1.5em}
    \label{fig:qualitative_result}
\end{figure}

\subsection{Asymmetric Traversal Strategy}

Rather than processing all $\mathcal{S}_{v,t}$ in a simple raster-scan order, we adopt an \emph{asymmetric traversal} strategy that balances temporal propagation and cross-view alignment.

Concretely, we iterate over camera indices $v = 0, \dots, V{-}2$, and for each $v$ we choose a different temporal traversal pattern:
\begin{itemize}
    \item For \textbf{even} camera indices (and the last camera pair $v = V{-}2$), we perform a \emph{full temporal sweep}, generating sub-grids $\mathcal{S}_{v,t}$ for all $t \in [0, T{-}2]$.
    This encourages strong temporal propagation along the time axis.
    \item For \textbf{odd} camera indices, we generate only $\mathcal{S}_{v,0}$ (i.e., using the first two time steps $t = 0, 1$).
    This enforces cross-view alignment between neighboring cameras while avoiding redundant temporal passes.
\end{itemize}

The resulting traversal order can be summarized as follows:
even-indexed camera rows perform dense temporal coverage, while odd-indexed rows act as cross-view bridges at the initial time step.
This pattern yields an overlapping chain of sub-grids that spans the entire $V \times T$ grid, ensuring that information fused by STGA in one region can be propagated to distant regions through CTP.


\subsection{Effect on STGA and CTP}

This asymmetric, overlapping traversal has two key effects:

\begin{itemize}
    \item \textbf{Local fusion via STGA.}  
    Within each $\mathcal{S}_{v,t}$, STGA jointly attends over adjacent views and neighboring time steps, producing locally coherent spatio-temporal features.
    Due to the overlapping structure, boundary frames participate in multiple sub-grids, implicitly coupling neighboring regions.

    \item \textbf{Global propagation via CTP.}  
    CTP operates along the traversal order $\Omega$, propagating tokens from $\mathcal{S}_{\text{prev}}$ to $\mathcal{S}_{\text{curr}}$ through inherited and flow-guided token replacement.
    Since the sub-grids overlap in both view and time, this propagation forms a connected path over the entire grid, enabling the fused information to spread globally while respecting camera–time structure.
\end{itemize}

Together, the asymmetric traversal and overlapping sub-grids provide a principled backbone for Grid-based Spatio-Temporal Propagation, ensuring that local STGA fusion and global CTP propagation jointly enforce consistent editing across all views and time steps.

\section{Vital Layer Range Analysis for STGA}
\label{supp:vital}

In the main paper, we apply Spatio-Temporal Sub-Grid Attention (STGA) only to a vital layer range of MM-DiT in order to enhance spatio-temporal consistency without overly harming editing fidelity.
Here, we provide a more detailed quantitative analysis of this design choice.

\paragraph{Experimental setup.}
We conduct a systematic study on the DyNeRF dataset using 3 scenes sampled at 1~FPS and 5 editing prompts per scene (15 sequences in total).
For each configuration, we enable STGA on a different continuous range of MM-DiT layers and keep all other components fixed.
We report (i) \textit{Warping Error}$\downarrow$~\cite{lai2018learning} for temporal consistency, (ii) \textit{MEt3R}$\downarrow$~\cite{asim2025met3r} for multi-view consistency, and (iii) \textit{CLIP Text-Image Directional Similarity}$\uparrow$ (CLIP$_{\text{dir}}$)~\cite{radford2021learning} for editing fidelity.

\paragraph{Effect of early-layer STGA.}
Without STGA, both Warping Error and MEt3R are the worst (46.37 and 1.22), indicating pronounced temporal flicker and cross-view inconsistency.
As we progressively introduce STGA from shallow layers (\(0\text{--}9\), \(0\text{--}19\), \(0\text{--}29\)), both consistency metrics steadily improve.
In particular, enabling STGA on the first 30 layers (0--29) yields a strong reduction in temporal and multi-view error (Warping Error 30.90, MEt3R 0.99), while preserving a relatively high CLIP$_{\text{dir}}$ score (0.088).
This configuration achieves the best overall trade-off: it significantly enhances spatio-temporal coherence compared to the baseline, yet maintains competitive editing fidelity.

\begin{table}[t]
    \centering
    \label{tab:vital_layer_supp}
    \vspace{0.4em}
    \resizebox{\linewidth}{!}{
    \begin{tabular}{lccc}
        \toprule
        \textbf{Layer Range} &
        Warp-Err $\scriptscriptstyle 10^{-3}$ $\downarrow$ &
        MEt3R $\scriptstyle 10^{-1}$  $\downarrow$ &
        CLIP$_{\text{dir}}$ $\uparrow$ \\
        \midrule
        W/o STGA     & 46.37 & 1.221 & 0.1111 \\
        0--9         & 32.54 & 1.022 & 0.1014 \\
        0--19        & 32.25 & 1.006 & 0.0946 \\
        \rowcolor{gray!15}
        0--29 & 30.90 & 0.993 & 0.0879 \\
        0--39        & 28.32 & 0.956 & 0.0468 \\
        25--35       & 37.47 & 1.578 & 0.0693 \\
        20--40       & 37.41 & 1.581 & 0.0683 \\
        15--45       & 28.96 & 1.321 & 0.0863 \\
        10--50       & 38.81 & 1.538 & 0.0661 \\
        49--59       & 41.06 & 1.270 & 0.0704 \\
        39--59       & 41.82 & 1.279 & 0.0703 \\
        29--59       & 39.49 & 1.250 & 0.0829 \\
        19--59       & 41.74 & 1.272 & 0.0694 \\
        All layers   & 41.25 & 1.265 & 0.0689 \\
        \bottomrule
    \end{tabular}}
    \caption{\textbf{Detailed vital layer range analysis for STGA.}
    This table reports the exact numerical values corresponding to the trend shown in Figure 3 of the main paper.}
    \vspace{-1.em}
\end{table}

\paragraph{Applying STGA too deep.}
Extending STGA too aggressively into deeper layers (e.g., 0--39 or mid-to-deep ranges such as 25--35, 20--40, 10--50, or 19--59) further reduces or even oscillates consistency metrics, but at the cost of a substantial drop in CLIP$_{\text{dir}}$.
For example, 0--39 attains the lowest Warping Error and MEt3R among all settings, but its CLIP$_{\text{dir}}$ score collapses to 0.047, indicating that over-attending within local spatio-temporal neighborhoods can oversmooth edits, weaken text alignment, and lead to texture repetition or view-dependent artifacts, as shown in Fig. 7 of the main paper.
Similarly, configurations that only activate STGA in deeper blocks (e.g., 39--59, All layers) neither recover the consistency of early-layer STGA nor preserve high editing fidelity, suggesting that late-stage modifications are less effective for enforcing stable geometry and motion.
\vspace{-1.em}

\begin{table*}[htbp]
\small
    \centering
    \renewcommand\arraystretch{1.0}
    \setlength{\tabcolsep}{1.5mm}{
    \begin{threeparttable}
    \begin{tabular}{ccccccccccc}
        \toprule
        & & \multicolumn{4}{c}{\textbf{2D Consistency}} 
        & \multicolumn{3}{c}{\textbf{Reconstruction Fidelity}} 
        & \multicolumn{2}{c}{\textbf{Editing Fidelity}} \\
        \cmidrule(r){3-6} \cmidrule(r){7-9} \cmidrule(r){10-11}
        \textbf{\footnotesize STGA} &
        \textbf{\footnotesize CTP} &
        \makecell{\footnotesize Local\\Warp-Err $\scriptscriptstyle 10^{-3}$} $\downarrow$ &
        \makecell{\footnotesize Global\\Warp-Err $\scriptscriptstyle 10^{-3}$} $\downarrow$ &
        \makecell{\footnotesize Local\\MEt3R $\scriptstyle 10^{-1}$}  $\downarrow$ &
        \makecell{\footnotesize Global\\MEt3R $\scriptstyle 10^{-1}$} $\downarrow$ &
        \makecell{\footnotesize PSNR} $\uparrow$ &
        \makecell{\footnotesize SSIM} $\uparrow$ &
        \makecell{\footnotesize LPIPS} $\downarrow$ &
        \makecell{\footnotesize CLIP$_{\text{dir}}$} $\uparrow$ &
        \makecell{\footnotesize CLIP$_{\text{sim}}$} $\uparrow$ \\
        \cmidrule(r){1-2} \cmidrule(r){3-6} \cmidrule(r){7-11}
         - & - & 56.98 & 58.37 & 1.0721 & 1.4312 & 26.14 & 0.7445 & 0.1408 & \underline{0.1930} & {0.5414} \\
        \checkmark & - & 38.64 & 42.33 & \underline{0.9277} & \underline{1.2953} & 28.08 & 0.7875 & \underline{0.1122} & {0.1872} & \underline{0.6407} \\
         - & \checkmark & \underline{29.44} & \underline{31.63} & 1.0695 & 1.4364 & \underline{28.74} & \underline{0.8013} & 0.1165 & \textbf{0.1944} & \textbf{0.6418}\\
        \cmidrule(r){1-2} \cmidrule(r){3-6} \cmidrule(r){7-11}
        \checkmark & \checkmark &  \textbf{28.94} & \textbf{30.69} & \textbf{0.9074} & \textbf{1.2657} & \textbf{29.25} & \textbf{0.8064} & \textbf{0.1006}  & 0.1849 & 0.6397 \\
        \bottomrule
    \end{tabular}
    \end{threeparttable}}
    \caption{\textbf{Ablation Study: Local and Global Consistency.} Our method improves both local and global 2D consistency, ensuring that each sub-grid remains coherent both internally (local) and with its neighbors (global). Each component, STGA and CTP, helps maintain temporal and multi-view consistency, improving overall 4D reconstruction. By enforcing a globally stable 4D structure, our method achieves more consistent spatio-temporal behavior and higher reconstruction fidelity. Although CLIP-based metrics~\cite{radford2021learning} show a slight drop due to the trade-off between semantic alignment and spatio-temporal coherence, our approach still delivers more stable and reliable 4D edits, avoiding the geometric and temporal artifacts seen in the ablated variants.}
    \vspace{-1.em}
    \label{tab:ablation}
\end{table*}

\begin{table*}[htbp]
\small
    \centering
    \renewcommand\arraystretch{1.0}
    \setlength{\tabcolsep}{1.0mm}{
    \begin{threeparttable}
    \begin{tabular}{ccccccccccc}
        \toprule
        & & \multicolumn{4}{c}{\textbf{2D Consistency}} 
        & \multicolumn{3}{c}{\textbf{Reconstruction Fidelity}} 
        & \multicolumn{2}{c}{\textbf{Editing Fidelity}} \\
        \cmidrule(r){3-6} \cmidrule(r){7-9} \cmidrule(r){10-11}
        \makecell{\footnotesize \textbf{CTP-Full}} &
        \makecell{\footnotesize \textbf{CTP-Flow}} &
        \makecell{\footnotesize Local\\Warp-Err $\scriptscriptstyle 10^{-3}$} $\downarrow$ &
        \makecell{\footnotesize Global\\Warp-Err $\scriptscriptstyle 10^{-3}$} $\downarrow$ &
        \makecell{\footnotesize Local\\MEt3R $\scriptstyle 10^{-1}$}  $\downarrow$ &
        \makecell{\footnotesize Global\\MEt3R $\scriptstyle 10^{-1}$} $\downarrow$ &
        \makecell{\footnotesize PSNR} $\uparrow$ &
        \makecell{\footnotesize SSIM} $\uparrow$ &
        \makecell{\footnotesize LPIPS} $\downarrow$ &
        \makecell{\footnotesize CLIP$_{\text{dir}}$} $\uparrow$ &
        \makecell{\footnotesize CLIP$_{\text{sim}}$} $\uparrow$ \\
        \cmidrule(r){1-2} \cmidrule(r){3-6} \cmidrule(r){7-11}
        - & - & 56.98 & 58.37 & 1.0721 & 1.4312 & 26.14 & 0.7445 & 0.1408 & \textbf{0.1930} & \textbf{0.6407} \\
        - & \checkmark & \underline{29.79} & \underline{32.66} & 0.9205 & 1.2813 & \underline{28.97} & \underline{0.7990} & \underline{0.1034} & 0.1852 & \underline{0.6402} \\
        \checkmark & - & 33.22 & 37.36 & \underline{0.9094} & \underline{1.2736} & 28.19 & 0.7906 & 0.1089 & \underline{0.1865} & 0.6400\\
        \cmidrule(r){1-2} \cmidrule(r){3-6} \cmidrule(r){7-11}
        \checkmark & \checkmark &  \textbf{28.94} & \textbf{30.69} & \textbf{0.9074} & \textbf{1.2657} & \textbf{29.25} & \textbf{0.8064} & \textbf{0.1006}  & 0.1849 & 0.6397 \\
        \bottomrule
    \end{tabular}
    \end{threeparttable}}
    \caption{\textbf{Ablation Study: Context Token Propagation (CTP).} This ablation study is conducted with STGA included to isolate the effect of CTP. The results show that our method maintains both local (within each sub-grid) and global consistency. Full Token Inheritance (CTP-Full) and Flow-Guided Token Replacement (CTP-Flow) play a crucial role in reinforcing temporal and multi-view coherence, enabling more accurate reconstruction of edited dynamic scenes. Although CLIP-based metrics~\cite{radford2021learning} show a slight trade-off, CTP significantly enhances spatio-temporal consistency and overall 4D editing fidelity.}
    \vspace{-1.em}
    \label{tab:ablation_token}
\end{table*}

\paragraph{Chosen configuration.}
Based on these observations, we adopt the 0--29 configuration as our default choice in the main paper.
This vital layer range provides a balanced compromise: it substantially improves temporal and multi-view consistency over the baseline and deep-only variants, while incurring only a modest decrease in CLIP$_{\text{dir}}$ relative to the no-STGA setting.
In practice, we find that this trade-off yields visually smoother 4D reconstructions and more reliable 4D scene editing, whereas configurations with either no STGA or overly deep STGA tend to produce flickering, geometric drift, or oversmoothed, weakly edited results.

\section{Additional Ablation Study}
\label{supp:ablation}

We conduct additional ablation study to evaluate the impact of Asymmetric Sub-Grid Traversal (AGT) in Dynamic-eDiTor.

\paragraph{Asymmetric Sub-Grid Traversal (AGT)}
For the ablation of AGT, we perform experiments without CTP, as it relies on the sliding mechanism introduced by AGT. AGT is designed to create overlapping regions between adjacent sub-grids, promoting smoother transitions and improving global consistency. We hypothesize that removing this overlap will yield results that remain locally consistent within each sub-grid but exhibit severe temporal and multi-view discontinuities across sub-grid boundaries.

To fairly assess global consistency, we introduce two new metrics: \textbf{Global Warping Error} and \textbf{Global MEt3R}.  Unlike their standard versions, these metrics are computed only between the left boundary frames of adjacent sub-grids, directly quantifying the discontinuities that AGT aims to mitigate. Additionally, we define per-frame consistency metrics as \textbf{Local Warping Error} and \textbf{Local MEt3R}, which are the original 2D consistency metric used in the main paper.

As shown in ~\cref{tab:ablation_agt}, removing AGT yields slightly higher local consistency because each sub-grid updates all its frames during every iteration. However, without any connection between sub-grids, clear discontinuities emerge between them, degrading overall 4D reconstruction quality. In contrast, AGT introduces overlap across sub-grids, enabling information to propagate between them and producing more coherent results, despite a slight reduction in local consistency, as only the non-overlapping frames are updated.
Overall, these findings confirm that AGT plays a crucial role in preserving global coherence, enabling higher-quality and more faithful 4D reconstruction than using STGA alone.

We also observe that the non-sliding variant yields slightly higher CLIP scores, reflecting the inherent trade-off between consistency and text alignment. Without sliding, the model can more aggressively edit each isolated sub-grid to match the prompt, but this comes at the cost of failing to produce a globally coherent 4D scene, which is the primary objective of our method.

\section{Monocular Video Setting}
\label{supp:mono}
4D dynamic scene editing typically refers to a multi-view video setting where sufficient spatio-temporal information is captured. However, to explore the applicability of Dynamic-eDiTor in a monocular setting, we evaluate our method on the DyCheck~\cite{gao2022dynamic} dataset using Deformable 3D Gaussian Splatting model~\cite{yang2023deformable3dgs}. Since a monocular dataset contains only a single camera, we modify the camera–time grid to a purely temporal grid:

{
\setlength{\abovedisplayskip}{5pt}
\setlength{\belowdisplayskip}{5pt}
\setlength{\abovedisplayshortskip}{5pt}
\setlength{\belowdisplayshortskip}{5pt}
\begin{equation}
Grid_{temp} = \{f_{t} \mid t \in [0, \dots, T] \},
\label{eq:Grid_mono}
\end{equation}
}

Accordingly, each sub-grid \(\mathcal{S}_{t}\) consists of consecutive frames along the temporal axis:

{
\begin{equation}
\mathcal{S}_{t} = \{ f_{t},\ f_{t+1},\ f_{t+2},\ f_{t+3} \}.
\label{eq:subgrid_mono}
\end{equation}
}

Based on this modified sub-grid, we apply same Spatio-Temporal Sub-Grid Attention (STGA) mechanism. However, because only temporally adjacent frames are available, the key and value sets \(K_{\mathcal{S}_{t}}\) and \(V_{\mathcal{S}_{t}}\) become:
{
\begin{equation}
\begin{aligned}
K_{\mathcal{S}_{t}} &= [K_{f_{t}}, K_{f_{t+1}}, K_{f_{t+2}}, K_{f_{t+3}}], \\
V_{\mathcal{S}_{t}} &= [V_{f_{t}}, V_{f_{t+1}}, V_{f_{t+2}}, V_{f_{t+3}}].
\end{aligned}
\label{eq:qkv_mono}
\end{equation}
}

Thus, STGA in the monocular setting becomes:
{
\begin{equation}
\begin{aligned}
\mathrm{STGA}&(\mathcal{S}_{t}) 
= \mathrm{softmax}\Big(
    [Q_{\text{txt}},\, \mathrm{RoPE}(Q_{f_{t}})] \: \cdot \\
&    [K_{\text{txt}},\, \mathrm{RoPE}(K_{\mathcal{S}_{t}})]^{\top} / \sqrt{d_k}
\Big)
\cdot [V_{\text{txt}},\, V_{\mathcal{S}_{t}}],
\label{eq:joint_attn_mono}
\end{aligned}
\end{equation}
}
where \(d_k\) denotes the dimensionality of the key vectors.

For Context Token Propagation (CTP), where the token representation is defined as \(\phi(\mathcal{S}_{t}) = \mathrm{STGA}(\mathcal{S}_{t})\), we employ two Context Token Propagation strategies: Full Token Inheritance and Flow-guided Token Replacement. Since the sub-grids overlap by two temporal frames, we directly replace the entire current token  \(\phi(\mathcal{S}_{curr})\) in these overlapped frames with the previous token  \(\phi(\mathcal{S}_{prev})\). For the non-overlapped region, which corresponds to the two rightmost frames of the sub-grid, we apply flow-guided token replacement. To propagate the most recent temporal information, we warp tokens from the rightmost frame of the overlapped region and replace the tokens of the two non-overlapping frames:
{
\setlength{\abovedisplayskip}{5pt}
\setlength{\belowdisplayskip}{5pt}
\setlength{\abovedisplayshortskip}{5pt}
\setlength{\belowdisplayshortskip}{5pt}
\begin{equation}
\hat{\phi}_{{\mathrm{r}}}(\mathcal{S}_{t}) \;=\;
\mathrm{Warp}\big(\mathbf{F}_{t \rightarrow t-1}(x,y),\, \phi_{{\mathrm{r}}}(\mathcal{S}_{t-1})\big),
\label{eq:token_warping_mono}
\end{equation}
}
where $\hat{\phi}_{\mathrm{r}}(\mathcal{S}_{t})$ denotes the warped tokens in the rightmost column of the patch and $\mathbf{F}_{t \rightarrow t-1}(x,y)$ represents the downsampled forward flow. 
To ensure precise replacement, we compute a validity mask M(x,y) and replace only tokens in valid regions with warped tokens.

As illustrated in ~\cref{fig:monocular}, Dynamic-eDiTor achieves stable and reliable monocular scene editing. Our model effectively maintains the temporal consistency, and both STGA and CTP contribute significantly to producing temporally coherent non-rigid appearance edits and semantic local editing.

\begin{figure}
    \centering
    \includegraphics[width=1\linewidth]{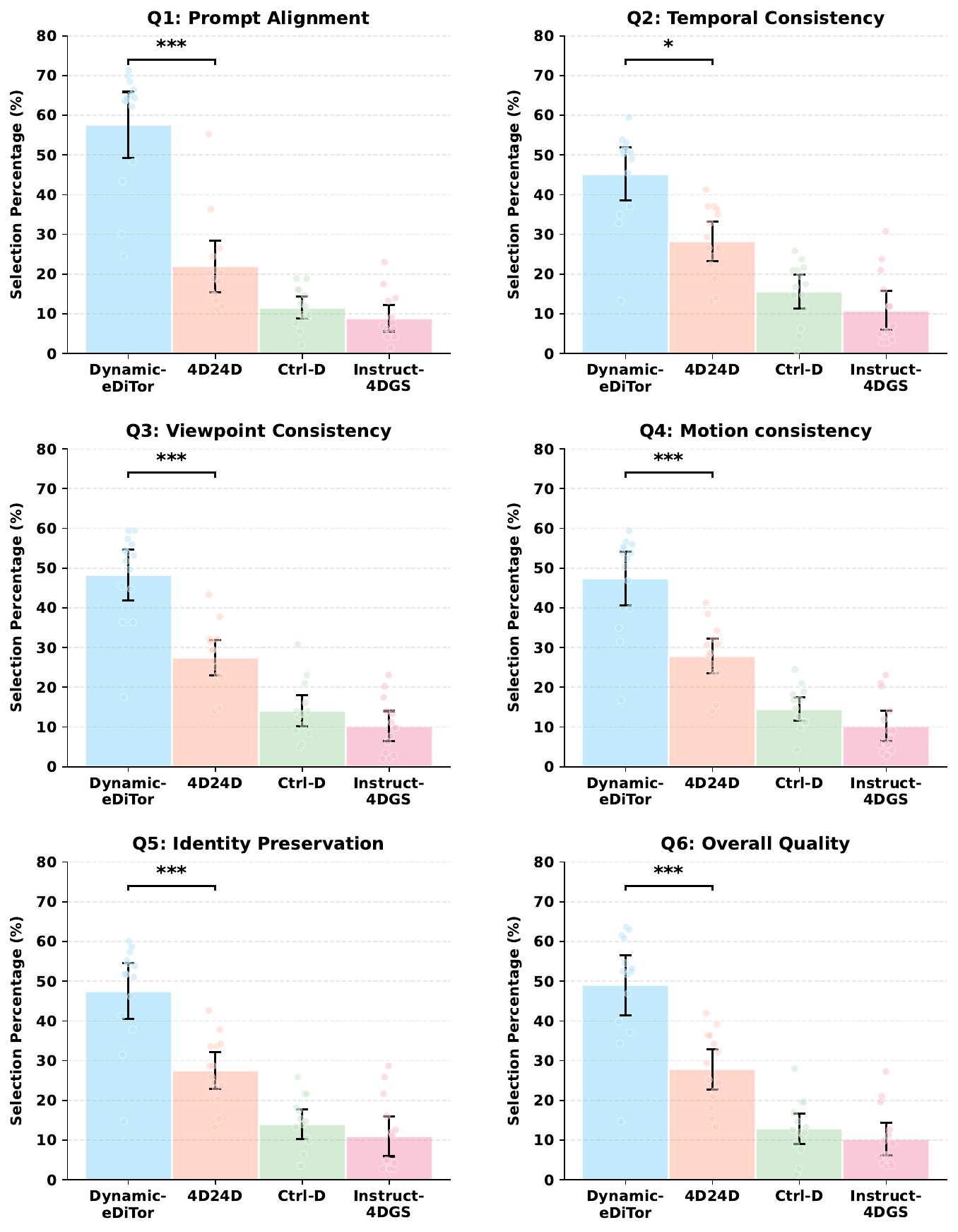}
    \caption{\textbf{User Study.} The user study results indicate a human preference for Dynamic-eDiTor, with superior ratings in both consistency and edited-quality categories compared to all baselines }
    \label{fig:usersutdy}
\end{figure}

\begin{figure*}[htp]
    \vspace{-2.2em}
    \centering
    \includegraphics[width=\linewidth]{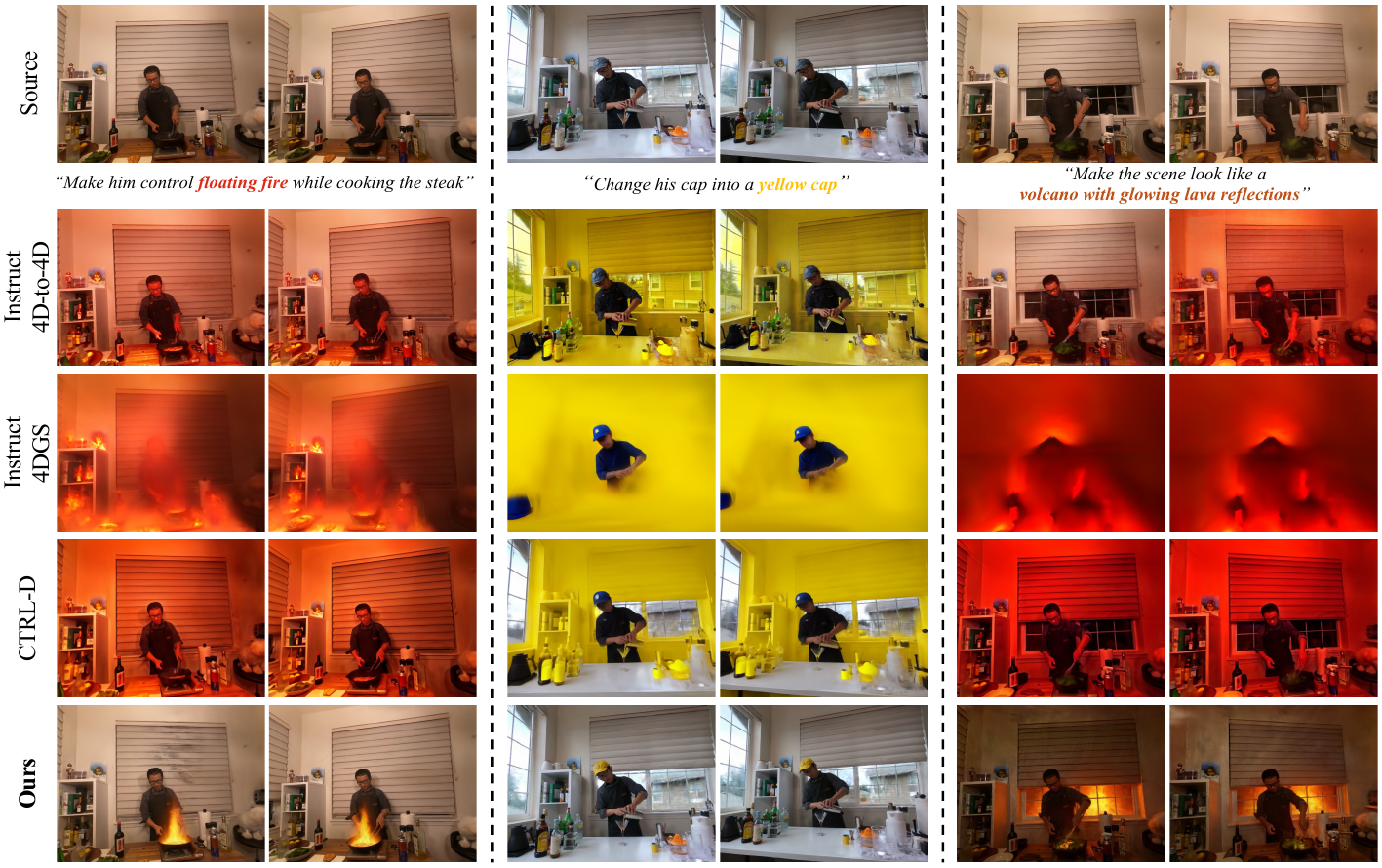}
    \vspace{-2.em}
    \caption{\textbf{Qualitative Results.} Dynamic-eDiTor enables higher-quality manipulation of non-rigid content and delivers more complete edits across the 4D scene. The upper row contains the original rendered frames, while the rows beneath show the edited 4DGS results from each baseline approach. Our method (bottom row) demonstrates superior correspondence to the text prompt and achieves strong edit fidelity while maintaining temporal and spatial consistency.} 
    \label{fig:supp_comparison}
\end{figure*}

\begin{figure*}[htp]
    \centering
    \includegraphics[width=\linewidth]{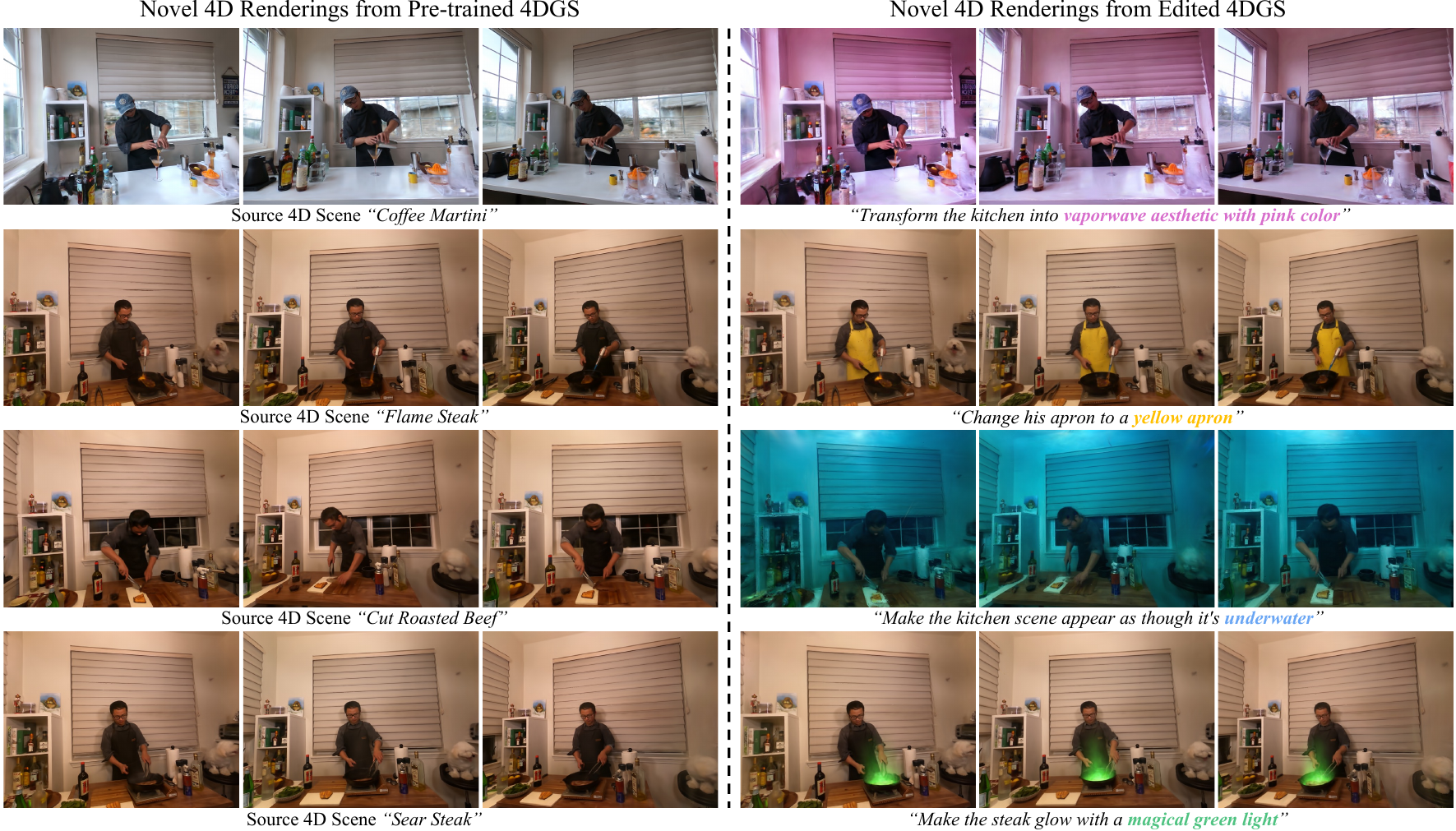}
    \vspace{-2.em}
    \caption{\textbf{Qualitative Results.} Dynamic-eDiTor preserves both multi-view and temporal consistency, enabling high-quality text-driven editing of pre-trained 4D Gaussian Splatting. It is capable of performing effective edits across diverse scenes as well as on local objects.} 
    \label{fig:supp_qualitative_result}
\end{figure*}

\begin{figure*}[htp]
    \centering
    \includegraphics[width=\linewidth]{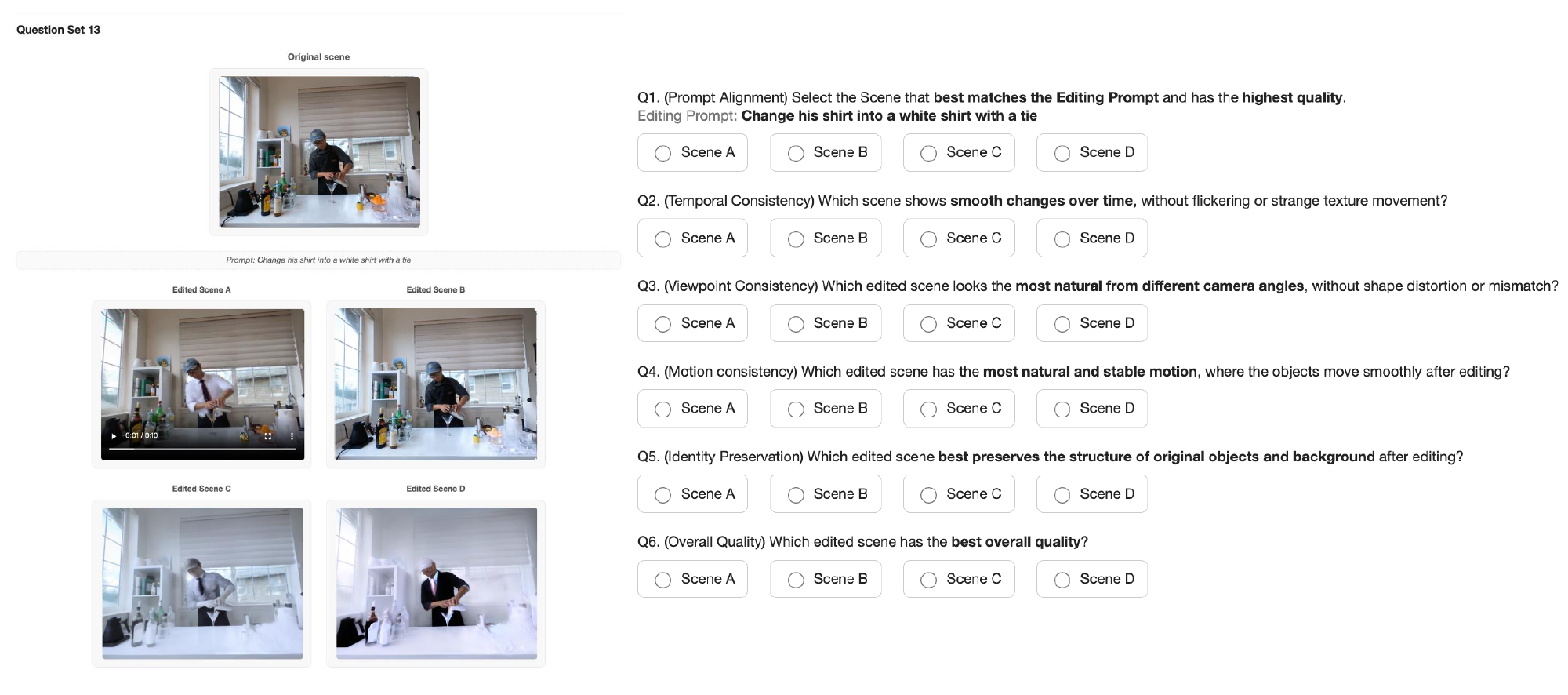}
    \vspace{-2.em}
    \caption{\textbf{User study interface and questionnaire.} We illustrate the interface used in our user study. Each participant is first shown the original scene and text prompt, then presented with four edited 4D-rendered video results (A–D) generated by different methods. Participants watch the videos and select the best method for each of the six evaluation criteria: prompt alignment (Q1), temporal consistency (Q2), viewpoint consistency (Q3), motion consistency (Q4), identity preservation (Q5), and overall quality (Q6).} 
    \label{fig:supp_userstudy_QA}
\end{figure*}

\end{document}